%% file: main.tex
\documentclass{article}
\PassOptionsToPackage{round}{natbib}

\usepackage[final]{neurips_2025}

\usepackage{hyperref}
\usepackage{url}
\usepackage{booktabs}
\usepackage{microtype}
\usepackage{lipsum}

\usepackage{wrapfig}
\usepackage{subcaption}
\usepackage{tikz}

\input{libs}

\theoremstyle{plain}
\newtheorem{theorem}{Theorem}[section]

\newtheorem{lemma}[theorem]{Lemma}

\theoremstyle{definition}
\newtheorem{definition}[theorem]{Definition}

\theoremstyle{remark}

\newtheorem*{theorem*}{Theorem}
\newtheorem*{lemma*}{Lemma}

\usepackage[capitalize]{cleveref}
\crefname{lemma}{Lemma}{Lemmas}
\crefname{definition}{Definition}{Definitions}

\title{Infinite Neural Operators:\\Gaussian processes on functions}

\author{%
  Daniel Augusto de~Souza\thanks{Corresponding author: \texttt{daniel.souza.21@ucl.ac.uk}}\\
  University College London
  \And
  Yuchen Zhu\\
  University College London
  \And
  Harry Jake Cunningham \\
  University College London
  \AND
  Yuri Saporito \\
  Fundação Getulio Vargas \\
  \And
  Diego Mesquita \\
  Fundação Getulio Vargas \\
  \And
  Marc Peter Deisenroth \\
  University College London
}

\begin{document}

\maketitle
\begin{abstract}
A variety of infinitely wide neural architectures (e.g., dense NNs, CNNs, and transformers) induce Gaussian process (GP) priors over their outputs.
These relationships provide both an accurate characterization of the prior predictive distribution and enable the use of GP machinery to improve the uncertainty quantification of deep neural networks.
In this work, we extend this connection to neural operators (NOs), a class of models designed to learn mappings between function spaces.
Specifically, we show conditions for when arbitrary-depth NOs with Gaussian-distributed convolution kernels converge to function-valued GPs.
Based on this result, we show how to compute the covariance functions of these NO-GPs for two NO parametrizations, including the popular Fourier neural operator (FNO).
With this, we compute the posteriors of these GPs in regression scenarios, including PDE solution operators.
This work is an important step towards uncovering the inductive biases of current FNO architectures and opens a path to incorporate novel inductive biases for use in kernel-based operator learning methods.
\end{abstract}

\input{text}

\begin{ack}
YZ acknowledges support by the Engineering and Physical Sciences Research Council with grant number EP/S021566/1.
YS was supported by Fundação Carlos Chagas Filho de Amparo à Pesquisa do Estado do Rio de Janeiro (FAPERJ) through the Jovem Cientista do Nosso Estado Program (E-26/201.375/2022 (272760)) and by Conselho Nacional de Desenvolvimento Científico e Tecnológico (CNPq) through the Productivity in Research Scholarship (306695/2021-9, 305159/2025-9).
DM was supported by the Fundação Carlos Chagas Filho de Amparo à Pesquisa do Estado do Rio de Janeiro (FAPERJ) (SEI-260003/000709/2023) and the Conselho Nacional de Desenvolvimento Científico e Tecnológico (CNPq) (404336/2023-0, 305692/2025-9).
\end{ack}

{\small
\bibliographystyle{plainnat}
\bibliography{main}
}
\newpage
\appendix
\input{supplement}
\end{document}

%% file: libs.tex
\usepackage{hyphenat}

\usepackage{ifthen}
\usepackage{amsmath,amsthm,amsfonts,amssymb,amscd,mathtools,bm,dsfont,empheq}
\allowdisplaybreaks

\usepackage{xcolor}

\ExplSyntaxOn
\cs_new_eq:NN \TextUppercase \text_uppercase:n
\cs_new_eq:NN \TextLowercase \text_lowercase:n
\ExplSyntaxOff

\renewcommand{\paragraph}{\par\textbf}

\definecolor{explain}{HTML}{ff7896}
\newcommand{\explain}[2][]{
\ifthenelse{ \equal{#2}{} }
{{\color{explain}\lipsum[#1][1-5]}}
{{\color{explain}\textbf{[#2]} \lipsum[#1][1-5]}}
}

\DeclareRobustCommand{\hexcircle}[2]{%
  \tikz[baseline=-0.7ex]{%
    \definecolor{myfill}{HTML}{#1}%
    \definecolor{mydraw}{HTML}{#2}%
    \draw[fill=myfill, draw=mydraw, line width=0.3ex] (0,0) circle (0.6ex);%
  }%
}
\DeclareRobustCommand{\hexdashcircle}[2]{%
  \tikz[baseline=-0.7ex]{%
    \definecolor{myfill}{HTML}{#1}%
    \definecolor{mydraw}{HTML}{#2}%
    \draw[fill=myfill, draw=mydraw, line width=0.3ex, dash pattern=on \pgflinewidth off 1pt] (0,0) circle (0.6ex);%
  }%
}
\DeclareRobustCommand{\hexline}[2]{%
  \tikz[baseline=0ex]{%
    \definecolor{myfill}{HTML}{#1}%
    \definecolor{mydraw}{HTML}{#2}%
    \draw[fill=myfill, draw=mydraw, line width=0.5ex, dash pattern=on \pgflinewidth off 1pt] (0,0.5ex) -- (1.3ex,0.5ex);%
  }%
}

\DeclarePairedDelimiter{\bracks}{[}{]}
\DeclarePairedDelimiterX{\bracksgiven}[2]{[}{]}{#1\:\delimsize|\:#2}
\DeclarePairedDelimiter{\braces}{\{}{\}}
\DeclarePairedDelimiterX{\setbuilder}[2]{\{}{\}}{#1\:\delimsize|\:#2}

\DeclarePairedDelimiter{\floor}{\lfloor}{\rfloor}
\DeclarePairedDelimiter{\parens}{\lparen}{\rparen}
\DeclarePairedDelimiter{\angles}{\langle}{\rangle}
\DeclarePairedDelimiterX{\parensgiven}[2]{\lparen}{\rparen}{#1\:\delimsize|\:#2}
\DeclarePairedDelimiterX{\iverson}[1]{[}{]}{\!\delimsize[#1\delimsize]\!}

\let\abs\det
\DeclarePairedDelimiter{\Verts}{\lVert}{\rVert}

\newcommand{\norm}[2][]{
  \ifthenelse{\equal{#1}{}}
  {\ensuremath{\Verts*{#2}}}
  {\ensuremath{\Verts*{#2}_{#1}}}
}

\DeclarePairedDelimiterX{\infdivx}[2]{\lparen}{\rparen}{#1\;\delimsize\|\;#2}

\newcommand{\GPdist}[1]{\operatorname{GP}\parens*{#1}}
\newcommand{\Normaldist}[2][]{
    \operatorname{\mathcal{N}}
    \ifthenelse{\equal{#1}{}}{\parens*{#2}}{\parensgiven*{#1}{#2}}
}
\DeclareMathOperator{\E}{\mathbb{E}}

\DeclareMathOperator{\cov}{cov}
\DeclareMathOperator{\Cov}{Cov}
\newcommand{\follows}{\sim}

\newcommand{\inprod}[2][]{
  \ifthenelse{ \equal{#1}{} }
  {\ensuremath{\angles*{#2}}}
  {\ensuremath{\angles*{#2}_{#1}}}
}

\let\vec\undefined
\newcommand{\blank}{\cdot}
\newcommand{\set}[1]{
\ifcat\noexpand#1\relax{#1}
\else\mathcal{\TextUppercase{#1}}\fi
}
\newcommand{\vec}[1]{\bm{\TextLowercase{#1}}}
\newcommand{\mat}[1]{\bm{\TextUppercase{#1}}}
\newcommand{\fun}[1]{\mathrm{#1}}
\newcommand{\vecfun}[1]{
\ifcat\noexpand#1\relax{\bm{#1}}
\else\mathbf{#1}\fi
}
\newcommand{\matfun}[1]{\mathbf{\TextUppercase{#1}}}
\newcommand{\op}[1]{{\TextUppercase{#1}}}

\newcommand{\call}[1]{\parens*{#1}}

\newcommand{\bcall}[1]{\bracks*{#1}}

\newcommand{\T}{\intercal}
\renewcommand{\d}{\mathrm{d}}

\let\L\undefined
\newcommand{\L}[1]{L^{#1}}

\newcommand{\Reals}[1][]{
  \ifthenelse{ \equal{#1}{} }
  {\ensuremath{\mathbb{R}}}
  {\ensuremath{\mathbb{R}^{#1}}}
}
\newcommand{\Nats}[1][]{
  \ifthenelse{ \equal{#1}{} }
  {\ensuremath{\mathbb{N}}}
  {\ensuremath{\mathbb{N}^{#1}}}
}
\newcommand{\Ints}[1][]{
  \ifthenelse{ \equal{#1}{} }
  {\ensuremath{\mathbb{Z}}}
  {\ensuremath{\mathbb{Z}^{#1}}}
}
\newcommand{\Complex}[1][]{
  \ifthenelse{ \equal{#1}{} }
  {\ensuremath{\mathbb{C}}}
  {\ensuremath{\mathbb{C}^{#1}}}
}
\newcommand{\Torus}[1][]{
  \ifthenelse{ \equal{#1}{} }
  {\ensuremath{\mathbb{T}}}
  {\ensuremath{\mathbb{T}^{#1}}}
}

\newcommand{\FT}{\operatorname{FS}}

\newcommand{\shortmathline}[2][]{
  \ifthenelse{ \equal{#1}{} }
  {\notag\mathrlap{#2}\hspace{0.5em}&\\*}
  {\notag\mathrlap{#2}\hspace{#1}&\\*}
}

\newcommand{\opname}{\operatorname}

%% file: text.tex
\section{Introduction}\label{sec:introduction}

Neural Operators~\citep[NOs,][]{Kovachki2023} are deep learning architectures designed to learn mappings between function spaces---with direct applications in many areas of science and engineering \citep{fourcastnet,pino}.
NOs generalize conventional convolutional neural networks using \emph{kernel integral operators}, which integrate the input function against a learnable kernel at each layer.
Importantly, unlike CNNs, NOs can be trained with inputs of mixed, arbitrary resolutions and output predictions in discretizations of arbitrary granularity.

Despite their growing adoption, most works on NOs are primarily empirical, and most of the theoretical properties of NOs are still unexplored.
In contrast, the convergence of Bayesian neural networks to Gaussian processes as their width goes to infinity has been amply studied~\citep{neal1995,Novak2019,Yang2019}.
However, due to the infinite dimensionality of function spaces, it is unclear whether GPs are a limiting case for NOs and, if this is the case, how to characterize them.

In this work, we elucidate this question and present a set of assumptions that guarantee the existence of the infinite limit of NOs as Gaussian elements in the space of operators.
Additionally, we present how to derive the covariance function for infinite-width NOs in an analogous fashion to the covariance functions of infinitely wide, densely connected NNs.
Finally, we characterize the infinite-width limit of Fourier neural operators (FNOs) and propose a novel Bayesian NO architecture based on Matérn GP-distributed integral kernels. \looseness=-1

Our experiments reinforce our theoretical results, showcasing the agreement between increasingly wide NOs and our derived expressions for the infinite limit at initialization.
Additionally, we compare the performance of these models in a regression setting.

\section{Background}\label{sec:background}

This section provides a brief background on NOs (\cref{subsec:operators}), along with basic notions of probability in Hilbert spaces (\cref{subsec:prob}) and Gaussian processes on functions. 

\subsection{Operator learning and neural operators}\label{subsec:operators}

\citet{Kovachki2023} propose neural operators, a family of parametrized operators.
Recall that multilayer perceptrons transform vectors using successive layers of sums of linear transformations followed by element-wise non-linear activation functions. Analogously, \citet{Kovachki2023} define the building layers of neural operators (NOs) as sums of both point-wise linear operations and kernel integral operators, possibly followed by point-wise element-wise non-linear activation functions.

Well-defined dot products in function spaces are central to coherently defining NOs. Thus, we will often assume functions lie in a vector space in which their dot product is finite wrt some measure $\mu_{\set{x}}$ over their domain $\set{x}$. We define
the Lebesgue space $\L{2}\call{\set{x},\mu_{\set{x}};\Reals[d]}$ as the equivalence classes of functions in this vector space that agree almost everywhere in $\set{x}$ with respect to $\mu_{\set{x}}$.
When clear from context, we will simply denote this vector space by $\L{2}\call{\set{x}}$.
Whenever needed to evaluate functions point-wise, we further assume the function lies in an appropriate Reproducing Kernel Hilbert Space (RKHSs). In this work, we will be using both the Lebesgue space $\L{2}\call{\set{x}}$ and RKHSs, when adequate. \looseness=-1

\textbf{Point-wise operators.}
These operations are carried over from standard neural networks.
Thus, given a function $\vecfun{f}\colon\set{X}\to\Reals[d]$, we consider dense layer-operations, with parameters $\mat{W}\in\Reals[b\times d]$, defined as $\parens{\mat{W}\vecfun{f}}\call{\vec{x}}\colon\set{X}\to\Reals[b] \coloneqq \mat{W}\vecfun{f}\call{\vec{x}}$, and element-wise activations, with a given $\fun{\sigma}\colon\Reals\to\Reals$, to be defined as ${\fun{\sigma}\bcall{\vecfun{f}}}_j\call{\vec{x}} \coloneqq \fun{\sigma}\call{\fun{f}_j\call{\vec{x}}}$.
By composing and adding results between layers, we can build neural operators that basically act just on the output of the functions.

\textbf{Kernel integral operator.}
\label{sec:kernel-int-op}
The majority of interesting behaviors require expanding the receptive field and aggregate results from different function evaluations into one.
The \emph{kernel integral operator} $\op{A}_{\matfun{k}}\colon\parens{\set{x}\to\Reals[d]}\to\parens{\set{y}\to\Reals[b]}$, parametrized by a matrix-valued kernel function $\matfun{K}: \set y \times \set x \to \Reals[b \times d]$ together with a measure $\mu_{\set x}$ on $\set{x}$, is defined as:
\begin{align}
    \op{A}_{\matfun{k}}\bcall{\vecfun{f}}\call{\vec{y}}
    \colon \set{y}\to\Reals[b]
    &=
    \int_{\set{x}}
    \matfun{k}\call{\vec{y},\vec{x}}\,
    \vecfun{f}\call{\vec{x}}
    \,\d\mu_{\set{x}}\call{\vec{x}}.
\end{align}
Under this operation, the function evaluated at a single evaluating point $\vec{y}$ linearly aggregates information on all evaluating points in the domain $\set{x}$ as modulated by the kernel $\matfun{k}$ and the measure $\mu_{\set{x}}$.
Note, that this function may not converge for all values of $\vec{y}$, but, for any kernel $\matfun{k}\in\L{2}\call{\set{y}\times\set{x},\mu_{\set y} \times \mu_{\set x}; \Reals[b\times d]}$, the operator $\op{A}_{\matfun{k}}\colon\L{2}\call{\set{x}, \mu_{\set x}}\to\L{2}\call{\set{y}, \mu_{\set Y}}$ is well-defined.

\textbf{Constructing neural operators.}
Given these building blocks, \citet{Kovachki2023} describe a neural operator as a three-part layered model. Firstly, a sequence of point-wise operators are applied to pre-process the function and change the dimension of its output. This is called the \emph{Lift layer}. The second component is a combination of point-wise and kernel integral operators, in the so-called \emph{Neural Operator layer}. Finally, the \emph{Projection layer}, a sequence of point-wise operators is applied to the final result.  \looseness=-1

Specifically, a neural operator layer combines a matrix-valued kernel $\matfun{K}$ and matrix $\mat{W}$ into \looseness=-1
\begin{align}
    \op{H}\bcall{\vecfun{f}}\call{\vec{x}}
    \colon
    \set{X}\to\Reals[h]
    =
    \op{A}_{\matfun{K}}\bcall{\vecfun{f}}\call{\vec{x}} +
    \mat{W}\vecfun{f}\call{\vec{x}}
    =
    \int_{\set{x}} \matfun{K}\call{\vec{x},\vec{z}}\,\vecfun{f}\call{\vec{z}}\,\d\mu_{\set{x}}\call{\vec{z}} +
    \mat{W}\vecfun{f}\call{\vec{x}}
    .
\end{align}
Setting the matrix-valued kernel to zero recovers the lift and projection layers. Therefore, a neural operator with depth $d$ and scalar output can be written succinctly as the composition:
\begin{equation}
    \op{Z}\bcall{\vecfun{f}}\call{\vec{x}}
    \colon
    \set{X}\to\Reals
    =
    \parens{
        \vec{w}^\T
        {}\circ{}
        \sigma {}\circ{}
        \op{H}_d
        \circ{}
        \cdots
        \sigma {}\circ{}
        \op{H}_1
    }
    \bcall{\vecfun{f}}
    \call{\vec{x}}
    .
\end{equation}

\subsection{Probability in Hilbert spaces}
\label{subsec:prob}

Given a probability space $\parens{\set{\Omega}, \Sigma, \fun{\mathbb{P}}}$,
and a Hilbert space $\set{H}$, random elements in $\set{H}$ are functions $x\colon\set{\Omega}\to\set{H}$, such that the inner product $\omega \in \Omega \mapsto \inprod{y,x\call{\omega}}_{\set{H}}$ is a real-valued random variable, for any $y\in\set{H}$.
As usual, we follow the standard notation of denoting the random elements/variables not as functions $x$ but as elements $x$.
Likewise, expectation is defined in terms of the random variables $\inprod{y,x\call{\omega}}$, for each $y\in\set{H}$.
We say that the expectation of $x$, when it exists, is the element of $\set{H}$, denoted by $\E\bcall{x}$, 
such that $\E\bcall{\inprod{y,x}} = \inprod{y, \E\bcall{x}}$, for any $y\in\set{H}$.

We denote the space of Hilbert--Schmidt (HS) operators mapping elements from a Hilbert space $\set{A}$ to $\set{B}$ by $\opname{HS}\call{\set{A};\set{B}}$. This space is the completion of the span of rank-one operators of the form ${a \otimes b} \colon \set{A} \to \set{B}$, defined as $(a \otimes b)(x) = \inprod{x, a}_{\set{A}}\, b$ for all $a \in \set{A}$ and $b \in \set{B}$.
For $\L{2}$ spaces, we have the isomorphism
\(
\opname{HS}\call{\L{2}\call{\set{X};\Reals[d]}, \L{2}\call{\set{Y};\Reals[b]}} \cong \L{2}\call{\set{X} \times \set{Y}; \Reals[{b \times d}]},
\)
under which 
\(
(\vecfun{f} \otimes \vecfun{g})\bcall{\vecfun{h}}\call\blank = \int_{\set{X}} \vecfun{g} \vecfun{f}^\T(\blank, \vec{x}) \vecfun{h}(\vec{x}) \, \d{\mu_{\set{x}}}(\vec{x}),
\)
where $\vecfun{f} \in \L{2}\call{\set{X};\Reals[d]}$, $\vecfun{g} \in \L{2}\call{\set{Y};\Reals[b]}$, and $\vecfun{g} \vecfun{f}^\T \in \L{2}\call{\set{X} \times \set{Y}; \Reals[{b \times d}]}$.

Moreover, the (cross-)covariance operator of two centered variables $x$ and $y$ is defined as the expectation of the tensor product $\E[x \otimes y]$. When this expectation exists, it is also a HS operator denoted as $\Cov\call{x,y}$.
From these definitions, we have that $\inprod{z_2, \Cov\call{x,y}\bcall{z_1}} = \cov\call{\inprod{z_2, y}\inprod{z_1, x}}$, for any $z_1,z_2\in\set{H}$. In $\L{2}$ spaces, we will make use of the isomorphism above and represent the covariance operator by its integration kernel. So, for any random elements $\vecfun{f}\in\L{2}\call{\set{X};\Reals[d]}$ and $\vecfun{g}\in\L{2}\call{\set{Y};\Reals[b]}$, we introduce the function
\(
\matfun{C}\bcall{\vecfun{f},\vecfun{g}}
\colon
\set{X}\times\set{Y}\to\Reals[b\times{d}]
\)
such that
\(
\Cov\call{\vecfun{f},\vecfun{g}}\bcall{\vecfun{h}}\call{\blank}
=
\int_{\set{x}}
    \matfun{C}\bcall{\vecfun{f},\vecfun{g}}\call{\blank,\vec{x}}\vecfun{h}\call{\vec{x}}
    \,\d\mu_{\set{X}}\call{\vec{x}}
\).

In this work, we will make use of an extension of the strong law of large numbers to random elements:
\begin{theorem}[Strong law of large numbers \citep{Mourier1956}]\label{thm:strong_law}
    Let $\set{H}$ be a separable Hilbert space and $\braces{x_j}_{j \in \mathbb{N}}$ be a countable sequence of identically distributed random elements. Consider the sample average $y_N = \parens{1/N}\sum_{j=1}^N x_j$. If, for any $j$, the expected norm $\E\bcall{\norm{x_j}}$ exists, then, the sequence $\braces{y_N}_{N \in \mathbb{N}}$ converges almost surely to the constant random element $y_\infty = \E\bcall{x_j}$.
\end{theorem}

\subsection{Operator valued kernels and Hilbert space valued Gaussian processes}

Now, given a set $\set{x}$ and a separable Hilbert space $\set{H}$, an \textit{operator-valued valued kernel} $\fun{C}\colon\set{x}\times\set{x}\to\opname{HS}\call{\set{H};\set{H}}$ is any Hermitian positive-definite function, i.e., for all $\vec{x},\vec{x}'\in\set{X}$, $\fun{C}\call{\vec{x},\vec{x}'} = \fun{C}\call{\vec{x}',\vec{x}}^\T$, and, for any $n>0$, $\braces{(\vec{x}_i, \vec{y}_i)}_{i=1}^n\subset\set{X}\times\set{H}$ and $\braces{\alpha_{ij}}_{i,j=1}^n\subset\Reals$, we have that $\sum_{i,j=1}^{n}\alpha_{ij}\inprod{\vec{y}_j, \fun{C}\call{\vec{x}_i,\vec{x}_j}\bcall{\vec{y}_i}} > 0$ \citep{Kadri2016}.

Consider an operator-valued kernel $\fun{C}\colon\set{X}\times\set{X}\to\opname{HS}\call{\set{H};\set{H}}$ such that $\vec{x} \mapsto\fun{C}\call{\vec{x}, \vec{x}}$ is of trace-class. We say $\fun{f}: \set{X} \times \Omega \to \set{H}$ is a centered Gaussian process with covariance function $\fun{C}$ if, for any $n > 0$ and $\braces{(\vec{x}_i, \vec{y}_i)}_{i=1}^n\subset\set{X}\times\set{H}$, the vector $\parens{ \inprod{y_1, \fun{f}\call{\vec{x}_1, \blank}}, \ldots,  \inprod{y_n, \fun{f}\call{\vec{x}_n, \blank}}}$ is a random element distributed as an $n$-dimensional Gaussian with covariance\looseness=-1
\begin{gather}
    \E\bcall{
        \inprod{y_2, \fun{f}\call{\vec{x}_2, \blank}}\ 
        \inprod{y_1, \fun{f}\call{\vec{x}_1, \blank}}
    }
    = \inprod{y_2,\fun{C}\call{\vec{x}_1,\vec{x}_2}\bcall{y_1}}.
\end{gather}
We denote this by $\fun{f}\follows\GPdist{0,\fun{C}}$. For simplicity, we also define $\fun{f}\call{\vec{x}} \coloneqq \fun{f}\call{\vec{x}, \blank}$.\looseness=-1

\section{Infinite-width neural operators as Gaussian processes}

It is well known that infinite-width limits of various Bayesian neural networks are Gaussian processes \citep{neal1995,Matthews2018}. We generalize this connection and show that infinite-width neural operators are function-valued Gaussian processes.

Analogous to \citet{Novak2019}, who place Gaussian priors on the convolution kernels of a CNN, the natural step towards function-valued GPs is to put independent GP priors on the component operators. Similarly, we require the weights and kernel for any component operator to be i.i.d. and with covariance shrinking with width. \autoref{theorem:inf-no} states the main result of this work. \looseness=-1

\begin{theorem}[Infinite-width neural operators are Gaussian processes]
    \label{theorem:inf-no}
    Let $\set{X}\subseteq\Reals[d_x]$ be a measurable space and let $\set{H}\call{\set{X};\Reals[J]}\subset \L{2}\call{\set{X};\Reals[J]}$ be an RKHS for any $J \in \Nats[+]$.
    Then, for a given depth $D\in\Nats[+]$, consider a vector of positive integers $\mat{J}=\bracks{J_0,J_1,\dotsc,J_{D-1},1}^\T\in\Nats[D+1]$ and a $\mat{J}$-indexed neural operators $\op{Z}_{\mat{J}}^{(D)}$ of depth $D$:
    \begin{equation}
    \op{Z}_{\mat{J}}^{(D)} \coloneqq
    \op{H}^{(D)} \circ \sigma \circ \op{Z}_{\mat{J}}^{(D-1)}
    \in \parens{\set{X}\to\Reals[J_0]} \to \parens{\set{X}\to\Reals},
    \end{equation}
    where,
    \begin{align}
    \op{Z}_{\mat{J}}^{(1)} &\coloneqq \op{H}^{(1)}
    \in\L{2}\call{\set{X};\Reals[J_{0}]} \to \set{H}\call{\set{X};\Reals[J_1]},\\
    \op{H}^{(\ell)} &\coloneqq (\op{A}_{\matfun{K}^{(\ell)}} + \mat{W}^{(\ell)})
    \in\L{2}\call{\set{X};\Reals[J_{\ell-1}]} \to \set{H}\call{\set{X};\Reals[J_\ell]},\\
    \mat{W}^{(\ell)}&\in\Reals[{J_\ell}\times{J_{\ell-1}}],\\
    \matfun{K}^{(\ell)}&\in\set{H}\call{
        \set{X}\times\set{X};
        \Reals[{J_\ell}\times{J_{\ell-1}}]
    },
    \end{align} 
    and \(
        \fun{\sigma}\colon\Reals\to\Reals
    \) such that $(\sigma\circ\fun{f})\in\L{2}\call{\set{x}}$ for any $\fun{f}\in\L{2}\call{\set{x}}$.
    
    When all parameters are independently distributed \emph{a priori} according to
    \begin{align}
        \mat{W}^{(\ell)} \follows \Normaldist{\vec{0},\sigma^2_{\ell}/{{J}_{\ell-1}}\mat{I}},
    \text{ and }
        \matfun{K}^{(\ell)}\follows\GPdist{\matfun{0}, \fun{c}_{\fun{k}^{(\ell)}}/{{J}_{\ell-1}}\mat{I}},
        \qquad \text{for }\ell\in\braces{1,\dotsc, d},
    \end{align}
    then, the iterated limit $\lim\limits_{J_{D-1}\to\infty}\cdots\lim\limits_{J_1\to\infty} \op{Z}_{\mat{J}}^{(D)}$, in the sense of \cref{def:iterated_convergence}, is equal to a function-valued GP ${\op{Z}_{\infty}^{(D)} \follows \GPdist{0, \fun{c}_{\infty}}}$, where $\fun{c}_{\infty}\bcall{\vecfun{f}, \vecfun{g}}$ is available in closed-form.
\end{theorem}
An outline of the proof is presented in \cref{subsec:proof-sketch}, where we present the explicit formula for $\fun{c}_{\infty}$, which depends on the conditional covariance function between layers. Before delving into these details, we introduce the compositionality property of covariance functions in \cref{subsec:covariance-functions}. This property enables the closed-form computation of the conditional covariances, thereby fully characterizing the limiting covariance function $\fun{c}_{\infty}$.

\subsection{Operator-valued covariance functions}\label{subsec:covariance-functions}
We realize the following crucial points: i) The covariance function only depends on the inner product of the values of the input functions, and ii) Using the strong law of large numbers, the covariance of the composition of operators can be described by composing its covariance functions. This is presented in the next lemma, with proof postponed to \cref{app:sec:composition}.

\begin{lemma}[Compositionality of covariance functions]\label[lemma]{lemma:composition}
Let $\op{B}_1\colon \L{2}\call{\set X; \Reals[d]} \to \L{2}\call{\set X; \Reals[J]}$ be a random operator and $\op{B}_2\colon \L{2}\call{\set X; \Reals[J]} \to \L{2}\call{\set X}$ be a centered function-valued Gaussian process.
If the following assumptions hold:
\begin{itemize}
    \item For all $\vecfun{f}\in\L{2}\call{\set{x};\Reals[d]}$ and $\vec{x}\in\set{X}$, each component of $\op{B}_1\bcall{\vecfun{f}}\call{\vec{x}}\in\Reals[J]$ is independent and identically distributed such that the covariance function $\matfun{c}_{\op{B}_1}\bcall{\vecfun{f},\vecfun{g}} = \fun{c}_{\op{B}_1}\bcall{\vecfun{f},\vecfun{g}}\mat{I}_{J}$;
    \item The covariance function of $\op{B}_2$ can be expressed, for all $\vecfun{f},\vecfun{g}\in\L{2}\call{\set{X};\Reals[J]}$ as $\fun{c}_{\op{B}_2}\bcall{\vecfun{f},\vecfun{g}} = \fun{c}_{\op{B}_2}\bcall{\frac{1}{J}\vecfun{g}^\T\vecfun{f}}$ and the function $\fun{h}\mapsto\fun{c}_{\op{B}_2}\bcall{\fun{h}}$ is a continuous map from $\L{2}\call{\set{x}\times\set{x}}$ to itself.
\end{itemize}
Then,
$\op{B}_2 \circ \op{B}_1$ converges in distribution to a function-valued Gaussian process as $J\to\infty$, and
\begin{align}
    \fun{c}_{\op{B}_2\circ\op{B}_1}\bcall{\vecfun{f}_1, \vecfun{f}_2} =
    \fun{c}_{\op{B}_2}\bcall{\fun{c}_{\op{B}_1}\bcall{\vecfun{f}_1, \vecfun{f}_2}}.
\end{align}
\end{lemma}

For each operator discussed in \cref{subsec:operators}, below we state the conditions under which they are function-valued Gaussian processes, and derive their covariance functions. 

\textbf{Point-wise linear operator.}
Given a vector $\vec{w}\in\Reals[d]$ and a function $\vecfun{f}\colon\set{x}\to\Reals[d]$, then, define the linear operator $\parens{\vec{w}^\T\vecfun{f}}\colon\set{x}\to\Reals$ such that $\parens{\vec{w}^\T\vecfun{f}}\call{\vec{x}} = \sum_{p=1}^{d}w_p\fun{f}_p\call{\vec{x}}.$
If the entries of the weight vector follow an i.i.d.\@ Gaussian distribution, i.e. $\vec{w}\follows\Normaldist{\vec{0},\sigma^2\mat{I}}$, then, this is a centered Gaussian process taking values from $\L{2}\call{\set{x};\Reals[d]}$ to $\L{2}\call{\set{x};\Reals}$ with covariance function:
\begin{align}
    \fun{c}_{\vec w}\bcall{\vecfun{f}_1, \vecfun{f}_2}\call{\vec{x}_1, \vec{x}_2} 
    &=  \sigma^2\,
        \vecfun{f}_{2}^\T\call{\vec{x}_2}\,
        \vecfun{f}_{1}\call{\vec{x}_1}
    ,\text{ such that},\\
    \cov\call{
        \inprod{\fun{h}_2, \vec{w}^\T\vecfun{f}_2}\,
        \inprod{\fun{h}_1, \vec{w}^\T\vecfun{f}_1}
    } &=
    \iint_{\set{x}}
        \fun{h}_2\call{\vec{x}_2}\,
        \fun{h}_1\call{\vec{x}_1}\,
        \fun{c}_{\vec{w}}\bcall{\vecfun{f}_1, \vecfun{f}_2}\call{\vec{x}_1,\vec{x}_2}
        \,\d\mu_{\set{x}}\call{\vec{x}_1}\d\mu_{\set{x}}\call{\vec{x}_2}
    .
\end{align}
Note that $\fun c_{\vec w}$ only depends on the function $\vecfun{f}_2^\T \vecfun{f}_1 : \set X \times \set X \to \Reals$, so we abuse notation and write $\fun{c}_{\vec w}\bcall{\vecfun{f}_1, \vecfun{f}_2} =
    \fun{c}_{\vec w}\bcall{\vecfun{f}_2^\T \vecfun{f}_1}$.
Moreover, this function is homogeneous: $\alpha\fun{c}_{\vec w}\bcall{\vecfun{f}_2^\T \vecfun{f}_1} = \fun{c}_{\vec w}\bcall{\alpha\vecfun{f}_2^\T \vecfun{f}_1}$, for $\alpha > 0$.

\textbf{Kernel integral operator.}
As defined in \cref{sec:kernel-int-op}, given a function $\vecfun{k}\colon\set{Y}\times\set{x}\to\Reals[d]$ and an input function $\vecfun{f}\colon\set{x}\to\Reals[d]$, we consider the linear operator $\op{A}_{\vecfun{k}^\T}\bcall{\vecfun{f}}\colon\set{Y}\to\Reals$.
If $\vecfun{k}$ follows an i.i.d.\@ GP such that $\vecfun{k}\in\L{2}\call{\set{Y}\times\set{x}}\follows\GPdist{0,\fun{c}_{\vecfun{k}}}$, then we have that $\op{A}_{\vecfun{k}^\T}$ is a centered function-valued GP mapping from $\L{2}\call{\set{x};\Reals[d]}$ to $\L{2}\call{\set{Y}}$ with covariance function, denoted here by:
\begin{align}
    \fun{c}_{\op{A}_{\vecfun{k}^\T}}\bcall{\vecfun{f}_1, \vecfun{f}_2}\call{\vec{y}_1, \vec{y}_2}
    &=  \iint_{\set{x}}
            \fun{c}_{\vecfun{k}}\call{\vec{y}_1,\vec{x}_1,\vec{y}_2,\vec{x}_2}\,
            \vecfun{f}_{2}^\T\call{\vec{x}_2}\,
            \vecfun{f}_{1}\call{\vec{x}_1}
        \,\d\mu_{\set{x}}\call{\vec{x}_1}\d\mu_{\set{x}}\call{\vec{x}_2}\\
    &= \op{A}_{\fun{c}_{\vecfun{k}}}\bcall{\vecfun{f}_2^\T\vecfun{f}_1}\call{\vec{y}_1,\vec{y}_2}
    .
\end{align}
Note $\fun{c}_{\op{A}_{\vecfun{k}^\T}}$ also only depends on the inner product of $\vecfun{f}_2$ and $\vecfun{f}_1$ and is homogeneous. Thus, we denote
\(\fun{c}_{\op{A}_{\vecfun{k}^\T}}\bcall{\vecfun{f}_1, \vecfun{f}_2} =
    \fun{c}_{\op{A}_{\vecfun{k}^\T}}\bcall{\vecfun{f}_2^\T \vecfun{f}_1}
\).
\looseness=-1

\textbf{Point-wise element-wise activation.}
Given a non-linear function $\fun{\sigma}\colon\Reals\to\Reals$, we abuse the notation and define the non-linear operator $\fun{\sigma}\bcall{\blank}\colon\L{2}\call{\set{X}}\to\L{2}\call{\set{X}}$ as
\begin{align}
    \fun{\sigma}\bcall{\fun{f}}\call{\vec{x}}
    = \fun{\sigma}\call{\fun{f}\call{\vec{x}}}.
\end{align}
Note that some restrictions on $\fun{\sigma}$ need to be placed for this to be a well-defined operator in $\L{2}\call{\set{x}}$. As an example of such condition, for their theoretical analysis, \citet{Kovachki2023} restricts activations to measurable linearly bounded functions, noting that the popular ReLU, ELU, tanh, and sigmoid activations satisfy this condition. In \cref{app:sec:activation}, we provide a proof that this condition is sufficient for finite measure domains.

Consider a centered Gaussian operator $\op{B}\colon\L{2}\call{\set{X}}\to\set{H}\call{\set{X}}$ with covariance function $\fun{c}_{\op{B}}$ such that $\set{H}\call{\set{X}}\subset\L{2}\call{\set{X}}$ is an RKHS with reproducing kernel $\fun{k}_{\set{H}}$.
When $\sigma\bcall{\blank}$ is a well-defined operator, the operator $\parens{\sigma\circ\op{B}}$ is a random operator in $\L{2}\call{\set{X}}\to\L{2}\call{\set{X}}$ with covariance function:
\begin{align}
    \fun{c}_{\parens{\sigma\circ\op{B}}}\call{\vec{x}_1, \vec{x}_2}
    &=
    \cov\call{
        \parens{\sigma\circ\op{B}}\bcall{\fun{f}_1}\call{\vec{x}_1},\,
        \parens{\sigma\circ\op{B}}\bcall{\fun{f}_2}\call{\vec{x}_2}
    }.
\end{align}
Now, since $\op{B}\bcall{\fun{f}_1}$ and $\op{B}\bcall{\fun{f}_2}$ are Gaussian processes with outputs in an RKHS $\set{h}$, we can consider the following bivariate Gaussian r.v. $\vec{b}_{[\fun{f}_1,\fun{f}_2]} = \bracks{\op{B}\bcall{\fun{f}_1}\call{\vec{x}_1}, \op{B}\bcall{\fun{f}_2}\call{\vec{x}_2}}^\T$:
\begin{align}
    \vec{b}_{[\fun{f}_1,\fun{f}_2]}
    \follows\Normaldist{
        \begin{bmatrix}0\\0\end{bmatrix},
        \begin{bmatrix}
            \fun{c}_{\op{B}}\bcall{\fun{f}_1, \fun{f}_1}\call{\vec{x}_1, \vec{x}_1} &
            \fun{c}_{\op{B}}\bcall{\fun{f}_1, \fun{f}_2}\call{\vec{x}_1, \vec{x}_2} \\
            \fun{c}_{\op{B}}\bcall{\fun{f}_2, \fun{f}_1}\call{\vec{x}_2, \vec{x}_1} &
            \fun{c}_{\op{B}}\bcall{\fun{f}_2, \fun{f}_2}\call{\vec{x}_2, \vec{x}_2} \\
        \end{bmatrix}
    }.
\end{align}
This random variable is well-defined due to the reproducing property, $\op{B}\bcall{\fun{f}}\call{\vec{x}} = \inprod{\fun{k}_{\set{H}}\call{\blank, \vec{x}}, \op{B}\bcall{\fun{f}}}$.

Thus, we can continue to conclude
\begin{align}
    \fun{c}_{\parens{\sigma\circ\op{B}}}\call{\vec{x}_1, \vec{x}_2}
    &=
    \int_{\Reals[2]}
        \sigma\call{{b}_{\fun{f}_1}}\,
        \sigma\call{{b}_{\fun{f}_2}}
    \ \Normaldist[\vec{b}_{[\fun{f}_1,\fun{f}_2]}]{\vec{0},{\mat{L}^\T\mat{L}}}
    \d\vec{b}_{[\fun{f}_1,\fun{f}_2]}
    \\&=
    \label{eq:activation-covariance}
    \int_{\Reals[2]}
        \sigma\call{\vec{l}_1^\T\vec{\xi}}\,
        \sigma\call{\vec{l}_2^\T\vec{\xi}}
    \ \Normaldist[\vec{\xi}]{\vec{0},{\mat{I}}}\d\vec{\xi}
    \\
    &\eqqcolon
    \fun{c}_{\sigma}\bcall{\fun{c}_{\op{B}}\bcall{\fun{f}_1,\fun{f}_2}}\call{\vec{x}_1,\vec{x}_2},
\end{align}
where
$\mat{L}$ is a square-root of the covariance matrix of $\vec{b}_{[\fun{f}_1,\fun{f}_2]}$ and $\vec{l}_i$ is the $i$-th row of this matrix.

The expected value $\fun{c}_{\sigma}$ as a function of $\vec{l}_1$ and $\vec{l}_2$ in \cref{eq:activation-covariance} is known as the \emph{dual kernel} of $\fun{\sigma}$. The dual kernels for many activation functions have closed-form solutions (e.g., sigmoid \citep[Eq. 10]{Williams1996} and ReLU \citep[Eq. 1]{Cho2009}) or can be efficiently approximated \citep{Han2022}. Any of these solutions can be directly used in our context by computing the covariance matrix of $\vec{b}_{[\fun{f}_1,\fun{f}_2]}$ and applying the rows of its square-root as arguments.

In conclusion, we construct an covariance function \(
    \fun{c}_{\sigma}\colon
    \set{H}\call{\set{X}\times\set{X}}\to
    \L{2}\call{\set{X}\times\set{X}}
\) such that, for a given covariance function \(
    \fun{c}_{\op{B}}\colon
    \L{2}\call{\set{X}\times\set{X}}\to
    \set{H}\call{\set{X}\times\set{X}}
\):
\begin{align}
    \inprod{
        \fun{h}_2\fun{h}_1,
        \fun{c}_{\sigma}\bcall{\fun{c}_{\op{B}}\bcall{\fun{f}_1,\fun{f}_2}}
    } &=
    \cov\call{
        \inprod{\fun{h}_1, \parens{\sigma\circ\op{B}}\bcall{\fun{f}_1}},
        \inprod{\fun{h}_2, \parens{\sigma\circ\op{B}}\bcall{\fun{f}_2}}
    },
\end{align}
for all $\fun{f}_1,\fun{f}_2,\fun{h}_1,\fun{h}_2\in\L{2}\call{\set{X}}$

\subsection{Outline of the proof for \cref{theorem:inf-no}}\label{subsec:proof-sketch}
    We now describe a sketch for the proof, we refer the readers to \cref{ap:proofs} for the complete proof.
    
    \textbf{Step 1.}
    We start by showing that, under the conditions of \cref{theorem:inf-no}, each linear layer in a neural operator layer is a function-valued Gaussian process when conditioned on its inputs.
    Moreover, as discussed in \cref{subsec:covariance-functions}, the conditional covariance function of each node on each layer only depends on the \emph{empirical covariance function} of its inputs \(
        \overline{\fun{c}}\bcall{\vecfun{f},\vecfun{g}}\call{\vec x', \vec x} = ({1}/{J})\sum_{j=1}^{J}
            \fun{g}_j\call{\vec{x}}\,
            \fun{f}_j\call{\vec{x}'}
    \). We denote this dependency by writing the conditional covariance function as $\fun{c}^{(\ell|\ell-1)}\bcall{\blank}\call{\vec{x},\vec{x}'}$.
    
    \textbf{Step 2.}
    Due to the chosen prior distribution of each layer, we know that each node in $\op{H}_{\ell}\bcall{\blank}\in\Reals[J_\ell]$ is i.i.d. and, therefore, we can apply Lemma~\ref{lemma:composition} to conclude that, as $J_{\ell-1}\to\infty$, the covariance \(
        \fun{c}^{(\ell|\ell-1)}\bracks{
            \op{H}_{\ell-1}\bcall{\vecfun{g}}^\T
            \op{H}_{\ell-1}\bcall{\vecfun{f}}
            /{J_{\ell-1}}
        }
    \) converges almost surely to \(
        \fun{c}^{(\ell|\ell-1)}\bracks{
            \fun{c}_{\op{H}_{\ell-1}}\bcall{\vecfun{f}, \vecfun{g}}
        }
    \).

    \textbf{Step 3.}
    Combining both steps, we show, by induction on $\ell$ up until $\ell=d$, that, as $J\to\infty$, the covariance function of $\op{Z}_J\bcall{\vecfun{f}}$ is simply the composition of all the previous covariances as denoted in Step 1.
    So, we have that the covariance function of $\op{Z}_\infty$ is:
    \begin{equation}
        \fun{c}^{(d)}\bcall{\vecfun{f}, \vecfun{g}} =
        \fun{c}^{(d|d-1)}[
        \fun{c}^{(d-1|d-2)}[
        \cdots
        \fun{c}^{(2|1)}\bracks{
        \fun{c}_{\op{H}_1}\bracks{\vecfun{f}, \vecfun{g}}
        }\cdots]
        .
    \end{equation}
    Finally, denote $\fun{c}^{(d)}$ as $\fun c_{\infty}$.

\section{Parametrizations and computations}\label{sec:para}

To apply the results of \Cref{theorem:inf-no}, we must specify a covariance function for the integral kernel operators $\op{A}_{\matfun{K}}$. This choice corresponds to selecting a particular neural operator parameterization, following the approach of \citet{Kovachki2023}.

In this section, we derive the operator-valued covariance functions for $\op{A}_{\mat{K}}$ under two parametrizations of the integral operator. The first is based on the band-limited Fourier Neural Operator (\Cref{sec:inf-fno}); the second models the kernel as a non-stationary process, with a prior distribution derived from the classical Matérn family of covariance functions (\Cref{sec:matern}).

A common assumption for both cases is that the input domain is compact. This ensures that samples of the kernel components $\vecfun{k}_j$ reside in a $\L{2}$ space. By further choosing the domain to be the $d_x$-dimensional flat torus $\Torus[d_x] = \Reals^{d_x} / 2\pi \Ints^{d_x}$, we are able to exploit Fourier analysis tools. In particular, by assuming that the input functions are band-limited enables tractable computations through the connection of Fourier series with discrete Fourier transforms for evaluations in regular grids.

\subsection{Fourier neural operator}\label{sec:fno}
Out of the parametrizations proposed by \citet{Kovachki2023}, the Fourier neural operator is the most popular due to its computational benefits. By imposing three assumptions into the convolutions kernel -- periodicity, shift-invariance, and band-limitedness -- we can use the convolution theorem  to compute the integrals using sums up to the chosen band-limit of the kernel in the Fourier space. \looseness=-1

Concretely,  assuming periodicity is equivalent to choosing the domain to be some $d_x$-dimensional flat torus $\set{x}=\Torus[d_x]$, and shift-invariance means kernels satisfy $\vecfun{k}_j \call{\vec{w}, \vec{x}} = \vecfun{k}_j \call{\vec{w} - \vec{x}}$, where we abuse notation and represent the kernel as a univariate function of the same name $\vecfun{k}_j : \Torus[d_x] \to \Reals[d]$. Under these conditions, any $\vecfun{k}_j$ admits a Fourier series representation:
\begin{align}
    \label{eq:kernel-fs}
    \vec{k}_j\call{\vec{w}-\vec{x}} = \sum_{\vec{s}\in\Ints[d_x]} \FT_{\vec s}\bcall{\vec{k}_j} \psi_{-\vec{s}}\call{\vec{w}-\vec{x}},
\end{align}
where $\FT_{\vec{s}}$ is the $\parens{s_1,\dotsc,s_{d_x}}$-th coefficient of the Fourier series and $\psi_{\vec{s}}\call{\vec{x}} = \exp\bcall{-i\cdot\vec{s}^\T\vec{x}}$, with $i=\sqrt{-1}$ being the imaginary unit. Moreover, to have a band-limited kernel implies that only finitely many Fourier coefficients are non-zero, i.e. there is some $B_j\in\Nats$, $1\leq j\leq d_x$, such that $\FT_{\vec{s}}\bcall{\vecfun{k}_j} = 0$, if $|s_j| > B_j$, for all $1\leq j\leq d_x$.

Under these conditions, despite all input functions $\vecfun{f}$ being represented with a (potentially infinite) Fourier series, by the convolution theorem, the NO layer $\op{H}_j\bcall{\vecfun{f}}$ is band-limited and its Fourier series coefficients can be computed directly from the product of Fourier coefficients of the kernel function $\vecfun{k}$ and the input function $\vecfun{f}$.
Thus, we have that:
\begin{align}
    \FT_{\vec{s}}\bcall{\op{H}_j\bcall{\vecfun{f}}}
    = \FT_{\vec s}\bcall{\vecfun{k}_j}^\T \FT_{\vec s}\bcall{\vecfun{f}} + {\vec{w}^{(1)}_j}^\T \FT_{\vec s}\bcall{\vecfun{f}}. \label{eq:ft-Akj}
\end{align}

\paragraph{Parameterization of an FNO.}
\label{sec:inf-fno}
Following \cref{subsec:covariance-functions}, when $\vecfun{k}$ is a $\Reals[d]$-valued GP, the kernel integral operator $\op{A}_{\vecfun k}$ is a function-valued Gaussian process with covariance function of $\op{A}_{\vecfun{k}}$ in terms of the covariance function of $\vecfun{k}$, $\matfun{C}_{\vecfun{k}}$:
\begin{align}
        \fun{c}_{\op{A}_{\vecfun{k}^\T}}\bcall{\vecfun{f}_1, \vecfun{f}_2}\call{\vec{z}_1, \vec{z}_2}
    &= \op{A}_{\matfun{c}_{\vecfun{k}}}\bcall{\vecfun{f}_2^\T\vecfun{f}_1}\call{\vec{z}_1,\vec{z}_2}.
\end{align}

The most popular choice proposed by \citet{Kovachki2023} is to directly parametrize the Fourier coefficients of the kernel. Thus, we let these $2B+1$ Fourier coefficients follow i.i.d. centered complex Gaussian distributions with variance $\sigma^2_{\vecfun{k}}$ (\cref{ap:sec:gp-bl}), obtaining the covariance function $\matfun{C}_{\vecfun{k}}$: \looseness=-1
\begin{align}
    \vecfun{k}\call{\vec z - \vec x} 
    &= \sum_{\mathclap{\vec{s}\in\braces*{-B,\dotsc,B}^{d_x}}}      
        \FT_{\vec s}\bcall{\vecfun{k}} \cdot \psi_{-\vec s}\call{\vec z - \vec x}
    \follows \GPdist{0, \matfun{C}_{\vecfun{k}}},\\
    \matfun{C}_{\vecfun{k}}\call{\parens*{\vec z - \vec x}, \parens*{\vec z' - \vec x'}} 
    &= 
        \sigma_{\vec k}^2 \mat{I}_{d}
        \sum_{\mathclap{\vec{s}\in\braces*{-B,\dotsc,B}^{d_x}}}
        \psi_{-\vec s}\call{\vec{z} -\vec x} \psi_{\vec s}\call{{\vec z'- \vec x'}}, 
\end{align}
where $B$ is a hyperparameter of the model controlling the band-limit of the integral kernel.

This allows us to derive a finite-sum representation of the covariance of $\op{A}_{\vecfun k}$ parameterized by $\sigma^2_{\vecfun{k}}$. 

\begin{align}
    \fun{c}_{\op{A}_{\vecfun{k}^\T}}
    \bcall{\vecfun{f}_1, \vecfun{f}_2}
    \call{\vec z, \vec z'}
&=  \sigma_{\vecfun k}^2
    \parens{2\pi}^{2d_x}\hspace{1ex}
        \sum_{\mathclap{\vec{s}\in\braces*{-B,\dotsc,B}^{d_x}}}
        \hspace{1ex}
        \FT_{-\vec{s}}\bcall{\vecfun{f}_2}^\T
        \FT_{\vec{s}}\bcall{\vecfun{f}_1}
        \psi_{-\vec{s}}\call{\vec{z}-\vec{z}'}
        \, \label{eq:fno-finite-form}.
\end{align}

\subsection{Toroidal Matérn operator}\label{sec:matern}
In this section, we propose a model in which the kernel does not admit a shift-invariant decomposition. Another popular decomposition used in the Gaussian process literature is the tensor-product factorization, where the covariance function of a GP factorizes over the input dimension. That is, $\fun{f}\colon\Reals[d_x]\to\Reals\follows\GPdist{0,\fun{c}}$, where $\fun{c}(\vec{a},\vec{b}) = \prod_j^{d_x} \fun{c}_j(a_j, b_j)$; although the covariance factorizes over the input dimensions, in general, samples from $\fun{f}$ do not.

Our proposal will make use of the ubiquitous Matérn family of covariance functions, which are characterized by the smoothness parameter $\nu$. Following \citet{slava2023}, we define the Matérn covariance functions in the $d_x$-dimensional flat torus $\Torus[d_x] = \Torus \otimes \cdots \otimes \Torus$ as:
\begin{align}
\fun{c}\call{\vec{x},\vec{x}'; \nu, \ell}
    &=  \parens{2\pi}^{-d_x}
        \sum_{\vec{n}\in\Ints^{d_x}}
        \fun{\psi}_{\vec{n}}\call{\vec{x}}\fun{\psi}_{-\vec{n}}\call{\vec{x}'}
        \hat{\fun{c}}\parens[\Big]{\sum_{j=1}^{d_x}n_j^2; \nu, \ell}.
\end{align}
where $\ell$ is the lengthscale hyper-parameter and the spectral density $\hat {\fun c}$ is defined as:
\begin{align}
    \hat{\fun{c}}\call{\lambda \ ; \nu, \ell} &= \begin{cases}
        \exp\bcall{-\frac{\ell^2}{2}\lambda},&
        \text{if $\nu=\infty$,}\\
        \parens*{\frac{2\nu}{\ell^2}+\lambda}^{-\nu-\frac{d}{2}},&\text{otherwise.}
    \end{cases}
\end{align}
In general, this kernel is not tensor-product factorized, but for the special case of $\nu=\infty$, the squared exponential covariance function, the factorization holds (Appendix~\ref{app:subsubsec:nu-infty-gives-product-kernel}). Thus, in general, we enforce the tensor-product factorization:
\begin{align}
\fun{c}\call{\vec{x},\vec{x}'; \nu, \vec \ell}
    &= \prod_{j=1}^{d_x} \fun{c}\call{x_j, x_j'; \nu, \ell_j}
    =  \parens{2\pi}^{-d_x}
        \sum_{\vec{n}\in\Ints^{d_x}}
        \fun{\psi}_{\vec{n}}\call{\vec{x}}\fun{\psi}_{-\vec{n}}\call{\vec{x}'}
        \prod_{j=1}^{d_x}\hat{\fun{c}}\call{n_j^2; \nu, \ \ell_j},
\end{align}
where $\vec{\ell}$ is the automatic relevance determination (ARD) lengthscale hyper-parameter.

\textbf{Parameterization of a toroidal Matérn operator.}
So, we consider a convolution kernel $\vec k: \set X \times \set X \to \Reals[d]$ defined as the product of Matérn covariance functions: 
\begin{align}
    \matfun c_{\vec k}\call{
    {\vec{z},\vec{x}, \vec{z}',\vec{x}'}
    } =
    \fun{c}\call{\vec{z},\vec{z}'; \nu_z, \ \vec\ell_z}\ 
    \fun{c}\call{\vec{x},\vec{x}'; \nu_x, \ \vec\ell_x}\
    \mat I_d. \label{eq:toroidal-kernel}
\end{align}
where $\fun{c}\call{\cdot, \cdot; \nu, \ \vec\ell}: \set x \times \set x \to \Reals$ is the Matérn covariance functions with smoothness parameter $\nu$ and length-scale $\vec\ell$.

Again, following \cref{subsec:covariance-functions}, we express the covariance of the operator as:
\begin{align}
        \fun{c}_{\op{A}_{\vecfun{k}^\T}}\bcall{\vecfun{f}_1, \vecfun{f}_2}\call{\vec{z}_1, \vec{z}_2}
    &= \op{A}_{\matfun{c}_{\vecfun{k}}}\bcall{\vecfun{f}_2^\T\vecfun{f}_1}\call{\vec{z}_1,\vec{z}_2}.
\end{align}

Thus, by using the identity \cref{eq:toroidal-kernel}, we can derive:
\begin{align}
    \fun{c}_{\op{A}_{\vec k^\T}}\bcall{\vecfun{ f}_1, \vecfun{f}_2}\call{\vec{z},\vec{z}'}
    &=  \fun{c}\call{\vec{z},\vec{z}'; \nu_z, \ \vec\ell_z}\ 
        \parens{2\pi}^{d_x}
        \sum_{\vec{n}\in\Ints^{d_x}}
        \FT_{\vec{n}}\bcall{\vecfun{f}_1}^\T
        \FT_{-\vec{n}}\bcall{\vecfun{f}_2}
        \prod_{j=1}^{d_x}\hat{\fun{c}}\call{n_j^2; \nu_x, \vec{\ell}_{x, j}}. \label{eq:toroidal-infinite-sum}
\end{align}

\section{Experimental validation}

In this section, we empirically show i) the agreement between finite width neural operators with increasing width and their corresponding infinite-width neural operator, and ii) evaluate our model against FNO in a regression task.

\cref{sec:empirical} explores the distribution of untrained randomly initialized Fourier neural operators of varying width and the distribution of the infinite-width FNO ($\infty$-FNO).
As expected from the theoretical results, these distributions should eventually match as the hidden dimension increases.

\cref{sec:regression}
considers two tasks: a synthetic regression example, where the task is to predict the output of a non-linear operator, and the task of predicting the final evolution of Burgers' equation given the initial state.
This situation is not covered in our theory, since it only applies to the distribution of the neural operators at initialization, but our experiments show the behavior of the posteriors of infinite-width neural operators against Adam trained finite-width neural operators of increasing width.

Throughout this section, our stating point is a single hidden-layer neural operator $\op{Z}\bcall{\fun{f}}\colon\Torus\to\Reals \coloneqq (\vec{w}_2^\T \circ \operatorname{ReLU} \circ{}\, \parens{\op{A}_{\matfun{K}} + \mat{W}_1})\bcall{\fun{f}}$. More details for each experiment can be found in \cref{ap:experiments}.
All experiments were implemented in Python, mainly based on the GPyTorch \citep{gpytorch} library, and run in a desktop computer using a Titan RTX. Code is avaliable at \url{https://github.com/spectraldani/infinite-neural-operator}.
\looseness=-1

\begin{figure}[t]
    \includegraphics[width=0.97\textwidth]{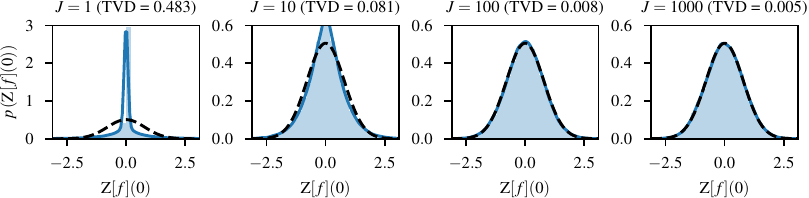}
    \caption{A density estimation of the \hexcircle{BBD5E8}{1F77B4} empirical distribution of the output of increasing channel dimension compared to the \hexdashcircle{FFFFFF}{000000} infinite width distribution. On top of each plot we show the total variation distance of the empirical distribution against the infinite width distribution.}
    \label{fig:fno_limit}
\end{figure}

\subsection{Empirical demonstration of results}\label{sec:empirical}

In this experiment, we demonstrate that our analytical computation of the variance for a neural operator layer $\op{H}$ agrees with empirical estimates, and we validate \cref{theorem:inf-no} by showing that the output of a neural operator $\op{Z}$ converges to a Gaussian distribution as the number of hidden channels $J$ increases. \looseness=-1

Throughout all experiments, the input function $\fun{f} \colon \Torus \to \Reals$ has band-limit $B = 3$, with its output values $\fun{f}\call{x}$ sampled from a uniform distribution $\mathcal{U}(-1,1)$. Both operators are evaluated at $x = 0$, so we analyze the empirical distributions of $\op{H}\bcall{\fun{f}}\call{0}$ and $\op{Z}\bcall{\fun{f}}\call{0}$, respectively.

\begin{wrapfigure}[13]{r}{0.4\textwidth}
    \vspace{-4mm}
    \centering
    \includegraphics[width=0.4\textwidth]{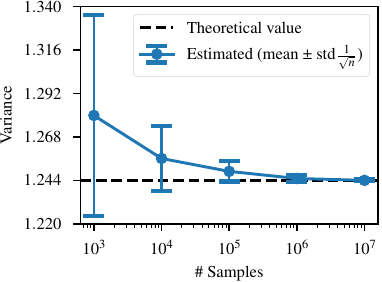}
    \vspace{-6mm}
    \caption{Plot of the MC estimate for the variance of $\op{H}\bcall{\fun{f}}\call{0}$ against our analytical computation (Sec.~\ref{subsec:covariance-functions}).}
    \label{fig:variance_estimate}
\end{wrapfigure}

Following \cref{sec:fno}, we parametrize the integral kernel operators using band-limited functions. The band-limit of the kernel is set equal to that of the input $\fun{f}$, and the kernel coefficients are drawn from a Gaussian distribution with unit variance scaled by the inverse of the number of hidden channels. \looseness=-1

As shown in \cref{fig:variance_estimate}, the empirical estimate of the variance converges to the theoretical value as the number of Monte Carlo samples increases, supporting the correctness of our variance computation. Furthermore, \cref{fig:fno_limit} shows that, as the number of hidden channels grows, the total variation distance (TVD) between the empirical distribution and a Gaussian distribution approaches zero, thereby further verifying the validity of \cref{theorem:inf-no}.

\subsection{Regression tasks}\label{sec:regression}

In this task, we're given $n$ pairs of 1D functions $\braces{\fun{f}_i, \fun{g}_i}_{i=1}^n$ evaluated in a grid with $m = 2B_m+1$ points. We consider FNOs of increasing width $J \in \Nats^+$, as well as $\infty$-FNOs, both with increasing kernel band-limits $B \in \braces{1, 5, 20}$. These models will be trained on two datasets: (a)~A operator generated by a randomly-initialized ground truth FNO $\op{Z}_{\opname{true}}$ with band-limit $B = 5$ and width $J = 1$. We sample $n = 100$ input functions $\fun{f}_i \colon \Torus \to \Reals$ with uniformily-distributed outputs $\mathcal{U}(-1, 1)$ and band-limit $B_m = 5$.
(b)~1D Burgers' equation dataset from \citet{PDEBench2022} with $\nu = 0.002$. The task is to predict the end state ($t=2$) given the initial condition ($t=0$). Due to memory constraints, we subsample the total dataset data to $n=100$ functions and a grid size of $m=103$.

\begin{figure}[t!]
    \centering
    \hspace{10mm}
    \begin{subfigure}[t]{0.4\textwidth}
        \centering
        \hspace{-13mm}
        \includegraphics[width=60mm]{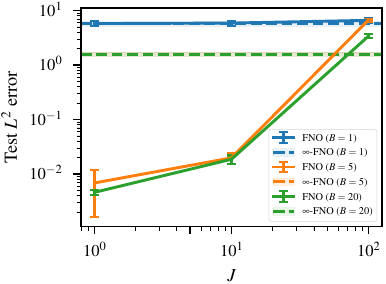}
        \caption{Synthetic data. $\infty$-FNO with $B=5$ (\hexline{FA842C}{FA842C}) and $B=20$ (\hexline{2E9F38}{2E9F38}) overlap.}
        \label{fig:regression}
    \end{subfigure}%
    \hfill
    \begin{subfigure}[t]{0.4\textwidth}
        \centering
        \hspace{-10mm}
        \includegraphics[width=60mm]{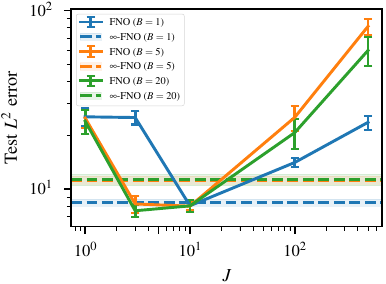}
        \caption{1D Burgers' equation ($\nu = 0.002$). $\infty$-FNO with $B=5$ (\hexline{FA842C}{FA842C}) and $B=20$ (\hexline{2E9F38}{2E9F38}) overlap.}
        \label{fig:burgers}
    \end{subfigure}
    \caption{Results for the regression experiments. Mean and std. of test $\L{2}$ loss as a function of width $J$ for different band-limits $B$.}
\end{figure}

The hyperparameters of the $\infty$-FNO are estimated using L-BFGS, while the parameters of the FNOs are optimized with Adam using a step size of $0.001$. We evaluate all models using 5-fold cross-validation and report the average and standard deviation of the empirical $\L{2}$ norm of the prediction error. For $\infty$-FNOs, we use the posterior mean as the prediction.

In general, we do not expect close agreement between the predictive performance of $\infty$-FNOs and finite-width FNOs, as the former corresponds to a Bayesian estimate while the latter are trained by minimizing an empirical risk, nonetheless, as observed in \cref{fig:regression,fig:burgers}, there is consistency between the gap of hyperparameters in the same model class.

In the synthetic case, as we know the band-limits of the ground truth operator, \cref{fig:regression} shows that the models are only able to accurately predict the output when their band-limits exceed that of the ground truth. 

\section{Related works}

\textbf{Infinite limits of stochastic NNs.}
The study of infinite-width Bayesian neural networks began with the seminal work of \citet{neal1995} and was later extended to deep architectures \citep{lee2018, Yang2019, Matthews2018}. Our analysis builds on the ideas developed by \citet{Matthews2018}. From the outset, these infinite-width models were considered "disappointing" \citep{neal1995}, a view reinforced by findings that neither the Bayesian limit nor the neural tangent kernel limit learns features from data \citep{Aitchison2020}. However, recent work shows these models still reflect the different inductive biases of their finite-width counterparts \citep{Novak2019}, and that alternative initialization distributions can enable feature learning in the infinite-width setting \citep{Yang2021}.

\textbf{Bayesian neural operators.}
Several works have investigated approximate Bayesian uncertainty quantification in finite-width neural operators using function-valued Gaussian processes.
\citet{magnani2022approximate, magnani2024} both employ last-layer Laplace approximations to construct GP approximations of the posterior distribution.
In addition, \citet{magnani2022approximate} considers the case where the kernel $\matfun{K}$ of the integral operator $\op{A}_{\matfun{K}}$ follows a Matérn GP prior.
However, their analysis is restricted to the finite-width regime on compact subsets of Euclidean space, whereas our work focuses on the flat torus. \looseness=-1

\textbf{Kernel methods for operator learning.}
\citet{batlle2024} propose the use of kernel methods for operator learning, leveraging operator-valued kernels and the representer theorem in their corresponding RKHS.
Their results are promising and highlight the potential of kernel-based approaches in this domain.
Our contribution introduces an additional way to construct operator-valued kernels based on neural operators, enabling new kernel-based models for operator learning.

\section{Discussion}

In this work, we formalized the concept of infinite-width Bayesian neural operators, established their existence (\cref{theorem:inf-no}), and described how to compute their associated covariance functions (\cref{sec:para}). We validated these results empirically (\cref{sec:empirical}) and further assessed the performance of these models in a regression setting (\cref{sec:regression}).

Our contributions lay a foundation for future investigations, particularly in bridging the gap between SGD-trained finite-width neural operators and their infinite-width counterparts. Addressing this challenge will require extending the neural tangent kernel framework \citep{jacot2018, lee2019} to settings involving Hilbert space-valued functions.
Moreover, while we focused on the ubiquitous FNO architecture, deriving covariance functions for other architectures, such as the graph neural operator \citep{Kovachki2023}, remains an open direction.

\textbf{Limitations.} Our current implementation for computing the required kernel quantities scales with cubically in both the evaluation grid size and the number of training functions. We anticipate that future work can improve computational efficiency by leveraging advances from the Gaussian process literature to improve scalability and efficiency \citep{slava2023, gilboa2013}.

%% file: supplement.tex
\setcounter{page}{1}
\title{Infinite Neural Operators:\\Gaussian processes on functions\\(Supplemental Materials)}
\makeatletter
\vbox{\hsize\textwidth
    \linewidth\hsize
    \vskip 0.1in
    \@toptitlebar
    \centering
    {\LARGE\bf \@title\par}
    \@bottomtitlebar}
    \vskip 0.3in \@minus 0.1in
\makeatother

\section{Covariance Function Computation}

In this section, we will work out in detail the computations of \cref{sec:para}, first for the Fourier neural operator (FNO) case and later for the toroidal Matérn operator.

\subsection{Fourier neural operator}\label{ap:sec:gp-bl}
Under the direct parametrization of the integral kernel operator, the coefficients of the kernel's Fourier series (FS) are parametrized and randomly sampled at initialization. Therefore, our first step is to derive what the Gaussian process distribution of a band-limited function with i.i.d.\@ Gaussian FS coefficients is.

\paragraph{Fourier series.}
Given a function on the $d_x$-dimensional torus $\vecfun{f}\call\blank\colon\Torus[d_x]\to\Reals[d]$, $\vecfun{f} = (\fun{f}_1,\ldots, \fun{f}_d)$, it can be represented in terms of a Fourier series:
\begin{align}
    \fun{f}_p\call{\vec{x}}
    &=  \sum_{\vec{s}\in\Ints[d_x]}
        \FT_{\vec{s}}\bcall{\fun{f}_p}\;
        \fun{\psi}_{-\vec{s}}\call{\vec{x}},\\
    \shortintertext{for $p\in\braces{1,\ldots,d}$, where,}
    \FT_{\vec{s}}\bcall{\fun{f}_p}
    &=  \parens{2\pi}^{-d_x}
        \int_{{\bracks{-\pi,\pi}^{d_x}}}
        \fun{f}_p\call{\vec{t}}\;
        \fun{\psi}_{\vec{s}}\call{\vec{t}}
        \,\d{\vec{t}},\\
    \fun{\psi}_{\vec{s}}\call{\vec{x}}
    &= \exp\bcall{-i\cdot\vec{s}^\T\vec{x}},
\end{align}
and $i=\sqrt{-1}$ is the imaginary unit.

Note that, as $f_p$ is a real-valued function, we also have that $\FT_{\vec{s}}\bcall{\fun{f}_p} = \overline{\FT_{-\vec{s}}}\bcall{\fun{f}_p}$, where $\Bar{z}$ is the complex conjugate.

\paragraph{Gaussian distributed band-limited functions.}
Consider the sequence $\hat{\vec{f}}\colon\bracks{-B,\dotsc,B}^{d_x}\to\Complex$ defined as:
\begin{align}
    \Re\hat{f}_{\vec{0}}
    &\follows \Normaldist{0,\sigma^2},
    &
    \Im\hat{f}_{\vec{0}}
    &= 0,\\
    \Re\hat{f}_{\vec{s}}
    &\follows \Normaldist{0,\sigma^2/2},
    &
    \Im\hat{f}_{\vec{s}}
    &\follows \Normaldist{0,\sigma^2/2},\\
    \Re\hat{f}_{-\vec{s}}
    &= \Re\hat{f}_{\vec{s}},
    &
    \Im\hat{f}_{-\vec{s}}
    &= -\Im\hat{f}_{\vec{s}},
\end{align}
where $\Re z$ and $\Im z$ are the real and imaginary parts of the complex number $z$, respectively, and all random variables are independent of each other, except the conjugate duals $\hat{f}_{\vec{s}}$ and $\hat{f}_{-\vec{s}}$. For $\vec{s}\neq\vec{0}$, the equations above can also be expressed as:
\begin{align}
    \begin{bmatrix}
        \Re\hat{f}_{\vec{s}}\\
        \Im\hat{f}_{\vec{s}}\\
        \Re\hat{f}_{-\vec{s}}\\
        \Im\hat{f}_{-\vec{s}}
    \end{bmatrix}
    = \mathcal{N}\call{
        \begin{bmatrix}
        0\\0\\0\\0
        \end{bmatrix},
        \frac{1}{2}
        \begin{bmatrix}
            \sigma^2&0&\sigma^2&0\\
            0&\sigma^2&0&-\sigma^2\\
            \sigma^2&0&\sigma^2&0\\
            0&-\sigma^2&0&\sigma^2\\
        \end{bmatrix}
    }.
\end{align}

With this in mind, the expectation of the product of two elements is:
\begin{align}
    \mathbb{E}\bracks{\hat{f}_{\vec{s}}\cdot{\hat{f}_{\vec{s}'}}}
    &=  \mathbb{E}\bracks{\Re\hat{f}_{\vec{s}} \Re\hat{f}_{\vec{s}'}}
        -\mathbb{E}\bracks{\Im\hat{f}_{\vec{s}} \Im\hat{f}_{\vec{s}'}}
        +i\mathbb{E}\bracks{\Re\hat{f}_{\vec{s}'} \Im\hat{f}_{\vec{s}}}
        +i\mathbb{E}\bracks{\Re\hat{f}_{\vec{s}} \Im\hat{f}_{\vec{s}'}}\\
    &=  \mathbb{E}\bracks{\Re\hat{f}_{\vec{s}} \Re\hat{f}_{\vec{s}'}}
        -\mathbb{E}\bracks{\Im\hat{f}_{\vec{s}} \Im\hat{f}_{\vec{s}'}}\\
    &= \begin{cases*}
        \mathbb{E}\bracks{\Re\hat{f}_{\vec{s}} \Re\hat{f}_{\vec{s}}} - \mathbb{E}\bracks{\Im\hat{f}_{\vec{s}} \Im\hat{f}_{\vec{s}}} & if $\vec{s}'=\vec{s}$,\\
        \mathbb{E}\bracks{\Re\hat{f}_{\vec{s}} \Re\hat{f}_{\vec{s}}} + \mathbb{E}\bracks{\Im\hat{f}_{\vec{s}} \Im\hat{f}_{\vec{s}}} & if $\vec{s}'=-\vec{s}$,\\
        0 & otherwise;
    \end{cases*}\\
    &= \begin{cases*}
        \sigma^2 & if $\vec{s}'=\vec{0}$ and $\vec{s}=\vec{0}$,\\
        \sigma^2/2 - \sigma^2/2 & if $\vec{s}'=\vec{s}$,\\
        \sigma^2/2 + \sigma^2/2 & if $\vec{s}'=-\vec{s}$,\\
        0 & otherwise;
    \end{cases*}\\
    &= \begin{cases*}
        \sigma^2 & if $\vec{s}'=-\vec{s}$,\\
        0 & otherwise.
    \end{cases*}
\end{align}

Thus, we can define the Gaussian process $\fun{f}\colon\Torus[d_x]\to\Reals$ through a Fourier series representation:
\begin{align}
    \fun{f}\call{\vec{x}}
    =   \hspace{1ex}
        \sum_{\mathclap{\vec{s}\in\braces{-B,\dotsc,B}^{d_x}}}
        \hspace{1ex}
        \hat{f}_{\vec{s}}\;
        \fun{\psi}_{-\vec{s}}\call{\vec{x}}.
\end{align}

We compute the covariance function of $\fun{f}$ as:
\begin{align}
    \fun{c}_{\fun{f}}\call{\vec{x}}
    &= \E\bcall{\fun{f}\call{\vec{x}}\cdot\fun{f}\call{\vec{x}'}}\\
    &= \E\bcall{
        \sum_{\vec{s}\in\braces{-B,\dotsc,B}^{d_x}}
        \hat{f}_{\vec{s}}\;
        \fun{\psi}_{-\vec{s}}\call{\vec{x}}
        \sum_{\vec{s}'\in\braces{-B,\dotsc,B}^{d_x}}
        \hat{f}_{\vec{s}'}\;
        \fun{\psi}_{-\vec{s}'}\call{\vec{x}'}
    }\\
    &=  \hspace{6ex}
        \sum_{\mathclap{\vec{s},\vec{s}'\in\braces{-B,\dotsc,B}^{d_x}}}
        \hspace{1ex}
        \mathbb{E}\bracks{
        \hat{f}_{\vec{s}}\;
        \fun{\psi}_{-\vec{s}}\call{\vec{x}}
        \;
        \hat{f}_{\vec{s}'}\;
        \fun{\psi}_{-\vec{s}'}\call{\vec{x}'}
    }\\
    &=  \hspace{6ex}
        \sum_{\mathclap{\vec{s},\vec{s}'\in\braces{-B,\dotsc,B}^{d_x}}}
        \hspace{1ex}
        \mathbb{E}\bracks{\hat{f}_{\vec{s}}\cdot\hat{f}_{\vec{s}'}}
        \fun{\psi}_{-\vec{s}}\call{\vec{x}}\;
        \fun{\psi}_{-\vec{s}'}\call{\vec{x}'}\\
    &=  \sigma^2
        \hspace{1ex}
        \sum_{\mathclap{\vec{s}\in\braces{-B,\dotsc,B}^{d_x}}}
        \hspace{1ex}
        \fun{\psi}_{-\vec{s}}\call{\vec{x}}\;
        \fun{\psi}_{\vec{s}}\call{\vec{x}'}\\
    &=  \sigma^2
        \hspace{1ex}
        \sum_{\mathclap{\vec{s}\in\braces{-B,\dotsc,B}^{d_x}}}
        \hspace{1ex}
        \fun{\psi}_{-\vec{s}}\call{\vec{x}-\vec{x}'}.
\end{align}

\subsubsection{Covariance after convolution $\fun{c}_{\op A_{\vecfun{k}^\T}}$}
Let us place a centered Gaussian distribution on the Fourier series of the band-limited convolution kernel $\vecfun{k}\colon\set{x}\times\set{x}\to\Reals[d]$, so that:
\begin{align}
    \matfun{c}_{\vecfun{k}}\call{
        \parens{\vec{z}-\vec{x}},
        \parens{\vec{z}'-\vec{x}'}
    }
    &=  \sigma^2\mat I_d
        \sum_{\mathclap{\vec{s}\in\braces*{-B,\dotsc,B}^{d_x}}}
            \hspace{1ex}
            \psi_{-\vec{s}}\call{
                \parens{\vec{z}-\vec{x}}-
                \parens{\vec{z}'-\vec{x}'}
            }\\
    &=  \sigma^2\mat I_d
        \sum_{\mathclap{\vec{s}\in\braces*{-B,\dotsc,B}^{d_x}}}
            \hspace{1ex}
            \psi_{-\vec{s}}\call{\vec{z}-\vec{x}}
            \psi_{\vec{s}}\call{\vec{z}'-\vec{x}'}\\
    &=  \sigma^2\mat I_d
        \sum_{\mathclap{\vec{s}\in\braces*{-B,\dotsc,B}^{d_x}}}
            \hspace{1ex}
            \psi_{-\vec{s}}\call{\vec{z}-\vec{z}'}
            \psi_{\vec{s}}\call{\vec{x}-\vec{x}'}
\end{align}

So, let us consider the quantity $\fun{c}_{\op A_{\vecfun k}}\bcall{\vecfun{f}_1, \vecfun{f}_2}$ for arbitrary functions $\vecfun{f}_1$ and $\vecfun{f}_2$:
\begin{align}
        \op{A}_{\fun{c}_{\vecfun{k}}}\bcall{\vecfun{f}_2^\T\vecfun{f}_1}
        \call{\vec{z},\vec{z}'}
&=
    \iint_{\mspace{8mu}\mathclap{\set{x}}}
        \vecfun{f}_2^\T\call{\vec{x}'}
        \matfun{c}_{\vecfun{k}}\call{
            \parens{\vec{z}-\vec{x}},
            \parens{\vec{z}'-\vec{x}'}
        }
        \vecfun{f}_1\call{\vec{x}}
    \,\d{\vec{x}}\d{\vec{x}'}
\\
&=
    \iint_{\mspace{8mu}\mathclap{\set{x}}}
        \vecfun{f}_2^\T\call{\vec{x}'}
        \parens[\Big]{
            \sigma^2\mat I_d
            \sum_{\mathclap{\vec{s}\in\braces*{-B,\dotsc,B}^{d_x}}}
            \hspace{1ex}
            \psi_{-\vec{s}}\call{\vec{z}-\vec{z}'}
            \psi_{\vec{s}}\call{\vec{x}-\vec{x}'}
        }
        \vecfun{f}_1\call{\vec{x}}
    \,\d{\vec{x}}\d{\vec{x}'}
\\
&=
    \sigma^2
    \sum_{\mathclap{\vec{s}\in\braces*{-B,\dotsc,B}^{d_x}}}
            \hspace{1ex}
            \psi_{-\vec{s}}\call{\vec{z}-\vec{z}'}
    \iint_{\mspace{8mu}\mathclap{\set{x}}}
        \psi_{\vec{s}}\call{\vec{x}-\vec{x}'}
        \vecfun{f}_2^\T\call{\vec{x}'}
        \vecfun{f}_1\call{\vec{x}}
    \,\d{\vec{x}}\d{\vec{x}'}
\\
&=
    \sigma^2
    \sum_{\mathclap{\vec{s}\in\braces*{-B,\dotsc,B}^{d_x}}}
            \hspace{1ex}
            \psi_{-\vec{s}}\call{\vec{z}-\vec{z}'}
    \parens{2\pi}^{2d_x}
    \FT_{[\vec{s},-\vec{s}]}\bcall{\vecfun{f}_2^\T\vecfun{f}_1}
\\
&=
    \sigma^2\parens{2\pi}^{2d_x}
    \sum_{\mathclap{\vec{s}\in\braces*{-B,\dotsc,B}^{d_x}}}
            \hspace{1ex}
            \psi_{-\vec{s}}\call{\vec{z}-\vec{z}'}
    \FT_{-\vec{s}}\bcall{\vecfun{f}_2}^\T
    \FT_{\vec{s}}\bcall{\vecfun{f}_1}.
\end{align}

\subsection{Toroidal Matérn operator}

\begin{definition}[Matérn family of kernels on a closed manifold]
    The Matérn family of kernels $\fun{c}$ with lengthscale $\ell$ in a $d$-dimensional closed manifold $\set{m}$ are described as:
\begin{align}
    \fun{c}\call{a,b; \nu, \ell} &=
    \sum_{k=1}^\infty
    \hat{\fun{c}}\call{\lambda_k; \nu, \ell}
    \cdot
    \fun{\phi}_k\call{a}\cdot\fun{\phi}_k\call{b},\\
    \hat{\fun{c}}\call{\lambda_k; \nu, \ell} 
    &= \begin{cases}
        \exp\bcall{-\frac{\ell^2}{2}\lambda_k}&
        \text{if $\nu=\infty$,}\\
        \parens*{\frac{2\nu}{\ell^2}+\lambda_k}^{-\nu-\frac{d}{2}}&\text{otherwise.}
    \end{cases},
\end{align}
where, $\lambda_k$ and $\fun{\phi}_k$ are the $k$-th eigenvalues and eigenfunctions, respectively, of the Laplace-Beltrami operator of the manifold $\set{m}$.
\end{definition}

For a 1-dimensional flat torus, an orthonormal eigensystem for the Laplace-Beltrami operator is:
\begin{align}
    \lambda_k &= \floor{k/2}^2;
    &
    \phi_k\call{x} &= \begin{cases*}
        {1}/{\sqrt{2}\pi} & if $k = 1$,\\
        \cos\call{\sqrt{\lambda_k}x}/\sqrt{\pi} & if $k = 2n$,\\
        \sin\call{\sqrt{\lambda_k}x}/\sqrt{\pi} & if $k = 2n+1$.
    \end{cases*}
\end{align}
Additionally, for a $d_x$-dimensional flat torus, an orthonormal eigensystem for the Laplace-Beltrami operator is given by the sum and product of the 1-dimensional eigensystem such that, given an index $\vec{k} = \bracks{k_1,\ldots,k_{d_x}}$, we have that $\lambda_{\vec{k}} = \sum_{j=1}^{d_x}\lambda_{k_j}$ and $\phi_{\vec{k}}\call{\vec{x}} = \prod_{j=1}^{d_x}\phi_{k_j}\call{x_j}$.

\paragraph{Expression of 1-dimensional toroidal kernel using complex exponentials.} For convenience, we will rewrite the series expansion of this kernel to use exponentials of complex numbers.

Start by noting that:
\begin{align}
    \phi_{2n}\call{a} \phi_{2n}\call{b}
    &= \frac{1}{\pi} \cos\call{n a} \cos\call{n  b},
    &
    \phi_{2n+1}\call{a} \phi_{2n+1}\call{b}
    &= \frac{1}{\pi} \sin\call{n a} \sin\call{n  b} ,
\end{align}
Therefore,
\begin{align}
    \phi_{2n}\call{a} \phi_{2n}\call{b} + \phi_{2n+1}\call{a} \phi_{2n+1}\call{b}
    &= \frac{1}{\pi} \cos\call{n a} \cos\call{n  b} + \frac{1}{\pi} \sin\call{n a} \sin\call{n  b} \\
    &= \frac{1}{\pi}\cos\call{n(a-b)}\\
    &= \frac{1}{2\pi} \parens*{
        \exp\bcall{in(a-b)}
        +
        \exp\bcall{-in(a-b)} 
    }\\
    &= \frac{1}{2\pi} \parens*{
        \exp\bcall{ina}
        \exp\bcall{-inb}
        +
        \exp\bcall{-ina}
        \exp\bcall{inb}
    }\\
    &= \frac{1}{2\pi} \parens*{
        \fun{\psi}_{-n}\call{a}
        {\fun{\psi}}_{n}\call{b}
        +
        \fun{\psi}_{n}\call{a}
        {\fun{\psi}}_{-n}\call{b}
    }
\end{align}
Now, we can rewrite the Matérn kernel expression as:
\begin{align}
    \fun{c}\call{a,b; \nu, \ell}
    &=
    \sum_{k=1}^\infty
    \hat{\fun{c}}\call{\lambda_k; \nu, \ell}
    \cdot
    \fun{\phi}_k\call{a}\cdot\fun{\phi}_k\call{b}\\
    &=
    \frac{1}{2\pi}\hat{\fun{c}}\call{\lambda_1; \nu, \ell} +
    \sum_{n=1}^\infty\hat{\fun{c}}\call{\lambda_{2n}; \nu, \ell}\parens*{
        \fun{\phi}_{2n}\call{a}\fun{\phi}_{2n}\call{b} +
        \fun{\phi}_{2n+1}\call{a}\fun{\phi}_{2n+1}\call{b}
    }\\
    &=
    \frac{1}{2\pi}\hat{\fun{c}}\call{0; \nu, \ell} +
    \sum_{n=1}^\infty\frac{1}{2\pi}\hat{\fun{c}}\call{n^2; \nu, \ell}\parens*{
        \fun{\psi}_{-n}\call{a}\fun{\psi}_{n}\call{b} +
        \fun{\psi}_{n}\call{a}\fun{\psi}_{-n}\call{b}
    }\\
    &=
    \frac{1}{2\pi}\hat{\fun{c}}\call{0; \nu, \ell} +
    \sum_{n=1}^\infty\frac{1}{2\pi}\hat{\fun{c}}\call{n^2; \nu, \ell}
        \fun{\psi}_{-n}\call{a}\fun{\psi}_{n}\call{b} +
    \sum_{n=1}^\infty\frac{1}{2\pi}\hat{\fun{c}}\call{n^2; \nu, \ell}
        \fun{\psi}_{n}\call{a}\fun{\psi}_{-n}\call{b}
    \\
    &=
    \frac{1}{2\pi}\hat{\fun{c}}\call{0; \nu, \ell} +
    \sum_{n=-1}^{-\infty}\frac{1}{2\pi}\hat{\fun{c}}\call{n^2; \nu, \ell}
        \fun{\psi}_{n}\call{a}\fun{\psi}_{-n}\call{b} +
    \sum_{n=1}^\infty\frac{1}{2\pi}\hat{\fun{c}}\call{n^2; \nu, \ell}
        \fun{\psi}_{n}\call{a}\fun{\psi}_{-n}\call{b}
    \\
    &=
    \frac{1}{2\pi}\hat{\fun{c}}\call{0}\fun{\psi}_{0}\call{a}\fun{\psi}_{-0}\call{b; \nu, \ell} +
    \sum_{n\in\Ints\setminus\braces{0}}\frac{1}{2\pi}\hat{\fun{c}}\call{n^2; \nu, \ell}
        \fun{\psi}_{n}\call{a}\fun{\psi}_{-n}\call{b}
    \\
    &=
    \frac{1}{2\pi}\sum_{n\in\Ints}\hat{\fun{c}}\call{n^2; \nu, \ell}
        \fun{\psi}_{n}\call{a}\fun{\psi}_{-n}\call{b}.
\end{align}

\paragraph{Product kernel for \texorpdfstring{$\Torus[d_x]$}{T^(d_x)}} In order to have one lengthscale per dimension, we will make a tensor product kernel where the kernel of a $d_x$-dimensional torus $\Torus[d_x]$ is the product of the 1-d toroidal kernel for each dimension:
\begin{align}
    \fun{c}\call{\vec{a},\vec{b}; \nu, \vec\ell_j}
    &=  \prod_{j=1}^{d_x}\fun{c}\call{a_j,b_j; \nu, \ell_j}\\
    &=  \prod_{j=1}^{d_x}\frac{1}{2\pi}\sum_{n\in\Ints}\hat{\fun{c}}\call{n^2; \nu, \ell_j}
        \fun{\psi}_{n}\call{a_j}\fun{\psi}_{-n}\call{b_j}\\
    &=  \parens{2\pi}^{-d_x}\prod_{j=1}^{d_x}\sum_{n\in\Ints}\hat{\fun{c}}\call{n^2; \nu, \ell_j}
        \fun{\psi}_{n}\call{a_j}\fun{\psi}_{-n}\call{b_j}\\
    &=  \parens{2\pi}^{-d_x}
        \sum_{\vec{n}\in\Ints^{d_x}}
        \prod_{j=1}^{d_x}\hat{\fun{c}}\call{n_j^2; \nu, \ell_j}
        \fun{\psi}_{n_j}\call{a_j}\fun{\psi}_{-n_j}\call{b_j}\\
    &=  \parens{2\pi}^{-d_x}
        \sum_{\vec{n}\in\Ints^{d_x}}
        \fun{\psi}_{\vec{n}}\call{\vec{a}}\fun{\psi}_{-\vec{n}}\call{\vec{b}}
        \prod_{j=1}^{d_x}\hat{\fun{c}}\call{n_j^2; \nu, \ell_j}.
\end{align}

\subsubsection{$\nu = \infty$ lets Matérn kernel be a product kernel}\label{app:subsubsec:nu-infty-gives-product-kernel}
Notice that when $\nu = \infty$ and $\ell_j = \ell$, we have 
\begin{align}
    \fun{c}\call{\vec{a},\vec{b}; \nu, \vec\ell}
    &=  \parens{2\pi}^{-d_x}
        \sum_{\vec{n}\in\Ints^{d_x}}
        \fun{\psi}_{\vec{n}}\call{\vec{a}}\fun{\psi}_{-\vec{n}}\call{\vec{b}}
        \prod_{j=1}^{d_x} \exp\bcall{-{\ell^2}n^2_j/2}.\\
    &=  \parens{2\pi}^{-d_x}
        \sum_{\vec{n}\in\Ints^{d_x}}
        \fun{\psi}_{\vec{n}}\call{\vec{a}}\fun{\psi}_{-\vec{n}}\call{\vec{b}}
         \exp\bcall{-\sum_{j=1}^{d_x}{\ell^2}n^2_j/2}.\\
    &=  \parens{2\pi}^{-d_x}
        \sum_{\vec{n}\in\Ints^{d_x}}
        \fun{\psi}_{\vec{n}}\call{\vec{a}}\fun{\psi}_{-\vec{n}}\call{\vec{b}}
         \exp\bcall{-\frac{\ell^2}{2}\sum_{j=1}^{d_x}{n^2_j}}.\\
    &=  \parens{2\pi}^{-d_x}
        \sum_{\vec{n}\in\Ints^{d_x}}
        \fun{\psi}_{\vec{n}}\call{\vec{a}}\fun{\psi}_{-\vec{n}}\call{\vec{b}}
         \hat{\fun{c}}\parens[\Big]{\sum_{j=1}^{d_x}{n^2_j}; \nu, \ell}.
\end{align}
With the proper rearrangement, we can see that this fits the definition of a Matérn kernel in the $\Torus[d_x]$, as the eigenvalues of its Beltrami-Laplace operator can be expressed as the sum of the eigenvalues for the 1-dimensional flat torus $\Torus$.

\subsubsection{Covariance after convolution $\fun{c}_{\op A_{\vecfun k}}$}
Let us place a centered factored Matérn prior in the convolution kernel $\vecfun{k}\colon\set{z}\times\set{x}\to\Reals[d]$, so that:
\begin{align}
    \matfun{c}_{\vecfun{k}}\call{
        {\vec{z},\vec{x}},
        {\vec{z}',\vec{x}'}
    } =
    \fun{c}\call{\vec{z},\vec{z}'; \nu_z, \vec\ell_z}\ 
    \fun{c}\call{\vec{x},\vec{x}'; \nu_x, \ell_x}
    \mat I_d.
\end{align}
When clear from context, we will suppress the dependency on the hyper-parameters of the Matérn kernel.

So, let us consider the quantity $\fun{c}_{\op A_{\vecfun k}}\bcall{\vecfun{f}_1, \vecfun{f}_2}$ for arbitrary functions $\vecfun{f}_1$ and $\vecfun{f}_2$:
\begin{align}
    \shortmathline[15mm]{
        \op{A}_{\fun{c}_{\vecfun{k}^\T}}\bcall{\vecfun{f}_2^\T\vecfun{f}_1}
        \call{\vec{z},\vec{z}'}
    }
    &=  \iint_{\set{x}}
        \vecfun{f}_2^\T\call{\vec x'}
        \matfun{c}_{\vecfun{k}}\call{
            {\vec{z},\vec{x}},
            {\vec{z}',\vec{x}'}
        }
        \vecfun{f}_1\call{\vec x}
        \,\d\vec{x}\d\vec{x}'
        \\
    &=  \iint_{\set{x}}
        \vecfun{f}_2^\T\call{\vec x'}
        \parens[\Big]{
            \fun{c}\call{\vec{z},\vec{z}'; \nu_z, \vec\ell_z}
            \fun{c}\call{\vec{x},\vec{x}'; \nu_x, \ell_x}
            \mat I_d
        }
        \vecfun{f}_1\call{\vec x}
        \,\d\vec{x}\d\vec{x}'
        \\
    &=
        \fun{c}\call{\vec{z},\vec{z}'}
        \iint_{\set{x}}
            \vecfun{f}_2^\T\call{\vec x'}\vecfun{f}_1\call{\vec x}
            \fun{c}\call{\vec{x},\vec{x}'}
        \,\d\vec{x}\d\vec{x}'
        \\
    &=
        \fun{c}\call{\vec{z},\vec{z}'}
        \sum_{\vec{n}\in\Ints^{d_x}}
        \parens{2\pi}^{-d_x}
        \iint_{\set{x}}
            \vecfun{f}_2^\T\call{\vec x'}\vecfun{f}_1\call{\vec x}
            \fun{\psi}_{\vec{n}}\call{\vec{a}}\fun{\psi}_{-\vec{n}}\call{\vec{b}}
        \,\d\vec{x}\d\vec{x}'
        \prod_{j=1}^{d_x}\hat{\fun{c}}\call{n_j^2; \nu, \ell_j}\\
    &=
        \fun{c}\call{\vec{z},\vec{z}'; \nu_z, \vec\ell_z}\ 
        \parens{2\pi}^{d_x}
        \sum_{\vec{n}\in\Ints^{d_x}}
        \FT_{\bracks{\vec{n},-\vec{n}}}\bcall{\vecfun{f}_2^\T\vecfun{f}_1}
        \prod_{j=1}^{d_x}\hat{\fun{c}}\call{n_j^2; \nu, \ell_j}\\
    &=
        \parens{2\pi}^{d_x}
        \fun{c}\call{\vec{z},\vec{z}'; \nu_z, \vec\ell_z}\ 
        \sum_{\vec{n}\in\Ints^{d_x}}
        \FT_{-\vec{n}}\bcall{\vecfun{f}_2}^\T
        \FT_{\vec{n}}\bcall{\vecfun{f}_1}
        \prod_{j=1}^{d_x}\hat{\fun{c}}\call{n_j^2; \nu, \ell_j}
\end{align}

\section{Proofs}

In this section, we include the proofs for \cref{lemma:composition}, \cref{theorem:inf-no}, and a short lemma on the well-defined-ness of the activation operator.

\subsection{Well-defined-ness of the point-wise element-wise activation operator}
\label{app:sec:activation}

\textbf{Lemma~B.1} {\itshape
Let $(\set{X}, \set{\Sigma}, \mu_{\set{X}})$ be a finite measure space, i.e. $\mu_{\set{X}}(\set{X}) \leq \infty$, and $\fun{\sigma}\colon\Reals\to\Reals$ a Borel measurable function such that
\begin{equation}
    \sup_{x\in\Reals} \frac{\abs{\fun{\sigma}\call{x}}}{1+\abs{x}} < C
\end{equation}
for some constant $C\in\Reals$. Then, the operator $\sigma\bcall{\fun{f}}\colon\L{2}\call{\set{X}}\to\L{2}\call{\set{X}} = \sigma\circ\fun{f}$ is well defined.
}
\begin{proof}
Remember that a function \(\fun{f}\) is in \(\L{2}\call{\set{X}}\) if, and only if,
\begin{equation}
\int_{\set X} \abs{\fun{f}\call{\vec{x}}}^2\ \d\mu_{\set{X}}\call{\vec{x}} < \infty.
\end{equation}

Now, from the linear boundedness condition, we know that for any $\fun{f}\in\L{2}\call{\set{X}}$ and any $\vec{x}\in\set{X}$:
\begin{align}
    \abs{\fun{\sigma}\call{\fun{f}\call{\vec{x}}}} &< C\parens{1+\abs{\fun{f}\call{\vec{x}}}},
\shortintertext{by squaring both sides and taking integrals,}
\int_{\set{x}}
\abs{\fun{\sigma}\call{\fun{f}\call{\vec{x}}}}^2
\ \d\mu_{\set{x}}\call{\vec{x}}
&<
C^2\int_{\set{x}}
\parens{1+\abs{\fun{f}\call{\vec{x}}}}^2
\ \d\mu_{\set{x}}\call{\vec{x}}
\end{align}

Now, note that the constant function \(1\) is in \(\L{2}\call{\set{X}}\) since $\int_{\set{X}} 1\ \d\mu_{\set{x}}\call{\vec{x}} = \mu_{\set{X}}\call{\set{X}} < \infty$ and that \(\abs{\fun{f}\call{\blank}}\) is in $\L{2}{\set{X}}$. Thus, from linearity, \(1+\abs{\fun{f}\call\blank}\) is also in $\L{2}{\set{X}}$ and $\int_{\set{x}} 1+\abs{\fun{f}\call{\vec{x}}}\,\d\mu_{\set{X}}\call{\vec{x}} < \infty$. Therefore,
\begin{align}
\int_{\set{x}}
\abs{\fun{\sigma}\call{\fun{f}\call{\vec{x}}}}^2
\ \d\mu_{\set{x}}\call{\vec{x}}
&<
\infty.
\end{align}
\end{proof}

\subsection{Compositionality of covariance functions}
\label{app:sec:composition}
\textbf{Lemma~\ref*{lemma:composition}.} {\itshape Let $\op{B}_1\colon \L{2}\call{\set X; \Reals[d]} \to \L{2}\call{\set X; \Reals[J]}$ be a random operator and $\op{B}_2\colon \L{2}\call{\set X; \Reals[J]} \to \L{2}\call{\set X}$ be a centered function-valued Gaussian process.
If the following assumptions hold:
\begin{itemize}
    \item For all $\vecfun{f}\in\L{2}\call{\set{x};\Reals[d]}$ and $\vec{x}\in\set{X}$, each component of $\op{B}_1\bcall{\vecfun{f}}\call{\vec{x}}\in\Reals[J]$ is independent and identically distributed such that the covariance function $\matfun{c}_{\op{B}_1}\bcall{\vecfun{f},\vecfun{g}} = \fun{c}_{\op{B}_1}\bcall{\vecfun{f},\vecfun{g}}\mat{I}_{J}$;
    \item The covariance function of $\op{B}_2$ can expressed, for all $\vecfun{f},\vecfun{g}\in\L{2}\call{\set{X};\Reals[J]}$ as $\fun{c}_{\op{B}_2}\bcall{\vecfun{f},\vecfun{g}} = \fun{c}_{\op{B}_2}\bcall{\frac{1}{J}\vecfun{g}^\T\vecfun{f}}$ and the function $\fun{h}\mapsto\fun{c}_{\op{B}_2}\bcall{\fun{h}}$ is a continuous map from $\L{2}\call{\set{x}\times\set{x}}$ to itself.
\end{itemize}
Then,
$\op{B}_2 \circ \op{B}_1$ converges in distribution to a function-valued Gaussian process as $J\to\infty$, and
\begin{align}
    \fun{c}_{\op{B}_2\circ\op{B}_1}\bcall{\vecfun{f}_1, \vecfun{f}_2} =
    \fun{c}_{\op{B}_2}\bcall{\fun{c}_{\op{B}_1}\bcall{\vecfun{f}_1, \vecfun{f}_2}}.
\end{align}}

\begin{proof}
    Consider a set of size $N\in\Nats[+]$, $\braces{(\vecfun{f}_n,\fun{h}_n)}_{n=1}^N\subset\L{2}\call{\set{x};\Reals[d]}\times\L{2}\call{\set{x}}$,
    then define the $N$-dimensional vector:
    \begin{align}
        \vec{z} \coloneqq
        \bracks*{
            \inprod{\fun{h}_1, \parens{\op{B}_2\circ\op{B}_1}\bcall{\vecfun{f}_1}},
            \dotsc,
            \inprod{\fun{h}_N, \parens{\op{B}_2\circ\op{B}_1}\bcall{\vecfun{f}_N}}
        }^\T
        \in\Reals[N].
    \end{align}
    Additionally, define the function:
    \begin{align}
        \bar{\fun{c}}_{\op{B}_1}\bcall{\vecfun{f}_j, \vecfun{f}_k}
        \colon
        \L{2}\call{\set{x};\Reals[d]}\times\L{2}\call{\set{x};\Reals[d]} \to
        \L{2}\call{\set{x}\times\set{x}}
        =
        \frac{1}{J}
        \op{B}_1\bcall{\vecfun{f}_k}^\T
        \op{B}_1\bcall{\vecfun{f}_j}.
    \end{align}

    Then, the conditional random variable $\vec{z}\mid \braces{\bar{\fun{c}}_{\op{B}_1}\bcall{\vecfun{f}_i, \vecfun{f}_j}}_{i,j=1}^{N}$ is Gaussian distributed with zero mean and covariance:
    \begin{align}
        \cov\parensgiven{z_i, z_j}{\bar{\fun{c}}_{\op{B}_1}\bcall{\vecfun{f}_i, \vecfun{f}_j}} = \inprod{
            \fun{h}_j, \inprod{
                \fun{h}_i,
                \fun{c}_{\op{B}_2}\bcall{
                    \bar{\fun{c}}_{\op{B}_1}\bcall{\vecfun{f}_i, \vecfun{f}_j}
                }
            }
        }.
    \end{align}

    We want to show that every $\vec{z}$ converges in distribution to a Gaussian distribution when $J\to\infty$, thus, it is useful to remember the following facts:
    \begin{itemize}
        \item \textbf{Multivariate Levy's continuity theorem.}
        A sequence of random variables $\braces{\vec{x}_j}_{j=1}^\infty$ converges to another one $\vec{x}_\infty$ if and only if the sequence of characteristic functions ${\phi_{\vec{x}_j}\call{\vec{t}} = \E\bcall{\exp\call{i\cdot \vec{t}^\T\vec{x}_j}}}$, where $i=\sqrt{-1}$, converges point-wise to $\phi_{\vec{x}_\infty}$.
        \item \textbf{Characteristic function of a $N$-dimensional Gaussian distribution.}
        If $\vec{x}\follows\Normaldist{\vec{0}, \mat{\Sigma}}$, then $\phi_{\vec{x}}\call{\vec{t}} = \exp\call{\vec{t}^\T\mat{\Sigma}\vec{t}}$ and $\phi_{\vec{x}}\call{\vec{t}} \leq 1$, for all $\vec{t}\in\Reals[N]$.
        \item \textbf{Strong law of large numbers.}
        As $J\to\infty$, the random element $\vecfun{K}\bcall{\vecfun{f}_i, \vecfun{f}_j}$ converges strongly to the constant $\fun{c}_{B_1}\bcall{\vecfun{f}_i, \vecfun{f}_j}$, for all $\vecfun{f}_i$.
        \item \textbf{Portmanteau theorem.} Given a sequence of random elements in $\set{H}$ converging in distribution $\braces{x_i}_{i=1}^\infty \to x_\infty$, then, $\lim_{i\to\infty}\E\bcall{f\call{x_i}} = \E\bcall{f\call{x_\infty}}$, for all bounded and continuous functions $\fun{f}\colon\set{H}\to\Reals$.
    \end{itemize}

    Thus, we begin with the characteristic function of the variable $\vec{z}$:
    \begin{align}
        \phi_{\vec{z}}\call{\vec{t}}
        &= \E\bcall{\exp\call{
            i\cdot\vec{t}^\T\vec{z}
        }},
        \shortintertext{by the tower rule, we can write}
        &= \E\bcall{\E\bracksgiven{\exp\call{
            i\cdot\vec{t}^\T\vec{z}
        }}{\braces{\bar{\fun{c}}_{\op{B}_1}\bcall{\vecfun{f}_j, \vecfun{f}_k}}_{j,k=1}^{N}}}
        = \E\bcall{\exp\call{
            \vec{t}^\T\mat{\Sigma}\vec{t}
        }},
    \end{align}
    where, $\bracks{\mat{\Sigma}}_{jk} = \cov\parensgiven{z_j, z_k}{\bar{\fun{c}}_{\op{B}_1}\bcall{\vecfun{f}_j, \vecfun{f}_k}}$ is a random variable.

    Now, because of the continuity of inner products and the assumption that $\fun{c}_{\op{B}_2}$ is continuous, we know that the mapping $\fun{h}\mapsto\cov\parensgiven{z_j, z_k}{\bar{\fun{c}}_{\op{B}_1}\bcall{\vecfun{f}_j, \vecfun{f}_k}=\fun{h}}$ is continuous. With this we take the limit:
    \begin{align}
        \lim_{J\to\infty}\phi_{\vec{z}}\call{\vec{t}}
        &= \lim_{J\to\infty}\E\bcall{\exp\call{
            \vec{t}^\T\mat{\Sigma}\vec{t}
        }};
        \shortintertext{using the portmanteau theorem, we get that:}
        \lim_{J\to\infty}\phi_{\vec{z}}\call{\vec{t}}
        &= \E\bcall{\exp\call{
            \vec{t}^\T\begin{bmatrix}
                \inprod{\fun{h}_1, \inprod{\fun{h}_1, \fun{c}_{\op{B}_2}\bcall{
                    \lim\limits_{J\to\infty}\bar{\fun{c}}_{\op{B}_1}\bcall{\vecfun{f}_1, \vecfun{f}_1}
                }}}&
                \cdots&
                \\
                \vdots& \ddots&\\
            \end{bmatrix}\vec{t}
        }},\\
        \shortintertext{and, finally, by the strong law of large numbers, we write the expectation as:}
        \lim_{J\to\infty}\phi_{\vec{z}}\call{\vec{t}}
        &= \exp\call{
            \vec{t}^\T\begin{bmatrix}
                \inprod{\fun{h}_1, \inprod{\fun{h}_1, \fun{c}_{\op{B}_2}\bcall{
                    \fun{c}_{\op{B}_1}\bcall{\vecfun{f}_1, \vecfun{f}_1}
                }}}&
                \cdots&
                \\
                \vdots& \ddots&\\
            \end{bmatrix}\vec{t}
        }.
    \end{align}
    Therefore, we have shown that $\vec{z}$ converges to a centered Gaussian distribution with:
    \begin{equation}
        \cov\call{z_j, z_k} =
        \cov\parensgiven{z_j, z_k}{
            \bar{\fun{c}}_{\op{B}_1}\bcall{\vecfun{f}_j, \vecfun{f}_k} = 
            \fun{c}_{\op{B}_1}\bcall{\vecfun{f}_j, \vecfun{f}_k}
        } =
        \inprod{\fun{h}_k, \inprod{\fun{h}_j, \fun{c}_{\op{B}_2}\bcall{
            \fun{c}_{\op{B}_1}\bcall{\vecfun{f}_j, \vecfun{f}_k}
        }}}
    \end{equation}
    Since the set $\braces{\vecfun{f}_n,\fun{h}_n}_{n=1}^N$ is arbitrary, we have shown that $\op{B}_2\circ\op{B}_1$ is a centered function-valued Gaussian process with covariance function $\fun{c}_{\op{B}_2}\bcall{\fun{c}_{\op{B}_1}\bcall{\blank, \blank}}$.
\end{proof}

\subsection{Infinite-width neural operators are Gaussian processes}
\label{ap:proofs}

\begin{definition}[Iterated convergence in distribution]\label[definition]{def:iterated_convergence}
    Let $X_{\vec{i}}$ be a random variable for each $\vec{i}=\bracks{i_1, \cdots, i_k}\subset \Nats[+]$. The iterated limit 
    \begin{align}
        \lim_{i_k\to\infty} \parens[\Big]{
            \lim_{i_{k-1}\to\infty} \parens[\Big]{
                \cdots
                \lim_{i_1\to\infty} \parens[\Big]{
                    X_{\vec{i}}
                }
                \cdots
            }
        }
    \end{align}
    whenever exists, is defined as the iterated limit in distribution. That is, suppose there are random variables $X_{\bracks{\infty, i_2, \cdots, i_k}}, X_{\bracks{\infty, \infty, \cdots, i_k}}, \cdots, X_{\bracks{\infty, \infty, \cdots, \infty}}$ such that for every $i_2, \cdots, i_k$
    \begin{align}
        X_{\bracks{i_1, i_2, i_3, \cdots, i_{k-1}, i_k} }
        &\overset{d}{\to} 
        X_{\bracks{\infty, i_2, i_3, \cdots, i_{k-1}, i_k}}
        \hspace{1cm}\text{as $i_1 \to \infty$}. \\
        X_{\bracks{\infty, i_2, i_3, \cdots, i_{k-1}, i_k}} 
        &\overset{d}{\to} 
            X_{\bracks{\infty, \infty, i_3, \cdots, i_{k-1}, i_k}} \\
        \notag&\vdots\\
        X_{\bracks{\infty, \infty, \infty, \cdots, \infty, i_k}} 
        &\overset{d}{\to} 
            X_{\bracks{\infty, \infty, \infty, \cdots, \infty, \infty}}
    \end{align}
    Then we define the iterated limit of $X_{\vec i}$ as 
    \begin{align}
        \lim_{i_k\to\infty} \parens[\Big]{
            \lim_{i_{k-1}\to\infty} \parens[\Big]{
                \cdots
                \lim_{i_1\to\infty} \parens[\Big]{
                    X_{\vec{i}}
                }
                \cdots
            }
        } &= X_{\bracks{\infty, \infty, \cdots, \infty}}
    \end{align}
\end{definition}

\textbf{Theorem~\ref*{theorem:inf-no}.} {\itshape
    Let $\set{X}\subseteq\Reals[d_x]$ be a measurable space and let $\set{H}\call{\set{X};\Reals[J]}\subset \L{2}\call{\set{X};\Reals[J]}$ be an RKHS for any $J \in \Nats[+]$.
    Then, for a given depth $D\in\Nats[+]$, consider a vector positive integers $\mat{J}=\bracks{J_0,J_1,\dotsc,J_{D-1},1}^\T\in\Nats[D+1]$ and a $\mat{J}$-indexed neural operators $\op{Z}_{\mat{J}}^{(D)}$ of depth $D$:
    \begin{equation}
    \op{Z}_{\mat{J}}^{(D)} \coloneqq
    \op{H}^{(D)} \circ \sigma \circ \op{Z}_{\mat{J}}^{(D-1)}
    \in \parens{\set{X}\to\Reals[J_0]} \to \parens{\set{X}\to\Reals},
    \end{equation}
    where,
    \begin{align}
    \op{Z}_{\mat{J}}^{(1)} &\coloneqq \op{H}^{(1)}
    \in\L{2}\call{\set{X};\Reals[J_{0}]} \to \set{H}\call{\set{X};\Reals[J_1]}\text{, and}\\
    \op{H}^{(\ell)} &\coloneqq (\op{A}_{\matfun{K}^{(\ell)}} + \mat{W}^{(\ell)})
    \in\L{2}\call{\set{X};\Reals[J_{\ell-1}]} \to \set{H}\call{\set{X};\Reals[J_\ell]},
    \end{align} 
    with \(
        \mat{W}^{(\ell)}\in\Reals[{J_\ell}\times{J_{\ell-1}}]
    \),
    and \(
        \matfun{K}^{(\ell)}\in\set{H}\call{
            \set{X}\times\set{X};
            \Reals[{J_\ell}\times{J_{\ell-1}}]
        }
    \).
    
    When all parameters are independently distributed \emph{a priori} according to
    \begin{align}
        \mat{W}^{(\ell)} \follows \Normaldist{\vec{0},\sigma^2_{\ell}/{{J}_{\ell-1}}\mat{I}},
    \text{ and }
        \matfun{K}^{(\ell)}\follows\GPdist{\matfun{0}, \fun{c}_{\fun{k}^{(\ell)}}/{{J}_{\ell-1}}\mat{I}},
        \qquad \text{for }\ell\in\braces{1,\dotsc, d},
    \end{align}
    then, the iterated limit $\lim\limits_{J_{D-1}\to\infty}\cdots\lim\limits_{J_1\to\infty} \op{Z}_{\mat{J}}^{(D)}$, in the sense of \cref{def:iterated_convergence}, is equal to a function-valued GP ${\op{Z}_{\infty}^{(D)} \follows \GPdist{0, \fun{c}_{\infty}}}$, where $\fun{c}_{\infty}\bcall{\vecfun{f}, \vecfun{g}}$ is available in closed-form.
}
\begin{proof}
    First, we note from \cref{subsec:covariance-functions} that the covariances \(
        \fun{c}_{\mat{W}^{(\ell)}}\bcall{\vecfun{f},\vecfun{g}}
    \) and \(
        \fun{c}_{\op{A}_{\matfun{k}^{(\ell)}}}\bcall{\vecfun{f},\vecfun{g}}
    \) are equal to:
    \begin{align}
        \matfun{c}_{\mat{W}^{(\ell)}}\bcall{\vecfun{f},\vecfun{g}}
        = \sigma^2_\ell\frac{1}{J_{\ell-1}}\vecfun{g}^\T\vecfun{f}\mat{I}_{J_\ell}\text{, and,}
        \matfun{c}_{\op{A}_{\matfun{k}^{(\ell)}}}\bcall{\vecfun{f},\vecfun{g}}
        = \op{A}_{\fun{c}_{\fun{k}^{(\ell)}}}\bcall{\frac{1}{J_{\ell-1}}\vecfun{g}^\T\vecfun{f}}\mat{I}_{J_\ell},
    \end{align}
    such that both depend on the empirical covariance \(
        \frac{1}{J_{\ell-1}}\vecfun{g}^\T\vecfun{f}
    \), for all $\vecfun{f},\vecfun{g}\in\L{2}\call{\set{X};\Reals[J_{\ell-1}]}$ and $\ell\in\Nats[+]$. Therefore, since $\op{H}^{(\ell)}$ is the sum of these two independent function-valued Gaussian processes, we have that $\op{H}^{(\ell)}\follows\GPdist{0, \fun{c}^{(\ell|\ell-1)}\mat{I}_{J_\ell}}$ such that:
    \begin{align}
        \fun{c}^{(\ell|\ell-1)}\bcall{\vecfun{f},\vecfun{g}} =
        \fun{c}^{(\ell|\ell-1)}\bcall{\vecfun{g}^\T\vecfun{f}/J_{\ell-1}} =
        \op{A}_{\fun{c}_{\fun{k}^{(\ell)}}}\bcall{
            \vecfun{g}^\T\vecfun{f}/J_{\ell-1}
        } +
        \sigma^2_\ell\vecfun{g}^\T\vecfun{f}/J_{\ell-1}
    \end{align}

    With this in mind, we proceed the proof by induction on the depth $D$.
    \paragraph{Base case.}
    For the base case $D=1$, we consider the operator $\op{Z}^{(1)}_{\mat{J}}$. Therefore, there are no limits to consider in this step. Nonetheless, as discussed in the previous paragraph, this quantity is a function-valued GP with covariance:
    \begin{align}
        \matfun{c}^{(1)}[\vecfun{f},\vecfun{g}] =
        \fun{c}^{(1|0)}\bcall{\vecfun{f},\vecfun{g}}\bm{I}_{J_1} = 
        \op{A}_{\fun{c}_{\fun{k}^{(1)}}}\bcall{
            \vecfun{g}^\T\vecfun{f}/J_{0}
        } +
        \sigma^2_1\vecfun{g}^\T\vecfun{f}/J_{0}.
    \end{align}
    Therefore, our claim is proven.
    \paragraph{Inductive step.}
    Our inductive hypothesis says that, for a specific $\ell\in\Nats[+]$, we have that the iterated limit $\lim\limits_{J_{\ell-1}\to\infty}\cdots\lim\limits_{J_1\to\infty} \op{Z}_{\mat{J}}^{(\ell)}$ converges in distribution to a ${\op{Z}_{\infty}^{(\ell)} \follows \GPdist{\vec{0}, \fun{c}^{(\ell)}\mat{I}_{J_{\ell}}}}$.

    As a first step, we would like to prove that
    \begin{gather}
        \lim\limits_{J_{\ell-1}\to\infty}\cdots\lim\limits_{J_1\to\infty}
        \op{H}^{(\ell+1)}\circ\sigma\circ\op{Z}_{\mat{J}}^{(\ell)}
    \shortintertext{converges in distribution to}
        \op{H}^{(\ell+1)}\circ\sigma\circ\op{Z}_{\infty}^{(\ell)}.
    \end{gather}
    Consider an arbitrary set of size $N\in\Nats[+]$,
    \begin{equation}
        (\set{f},\set{h}) = 
        \braces{
        (\vecfun{f}_1,\vecfun{h}_1),
        \dotsc,
        (\vecfun{f}_N,\vecfun{h}_N)
        }
        \subset\L{2}\call{\set{x};\Reals[J_0]}\times\L{2}\call{\set{x};\Reals[J_{\ell+1}]},
    \end{equation}
    and define the variables $\vec{z}\bcall{\set{f},\set{h}}\in\Reals[N]$ and $\matfun{Z}_{\mat{J}}^{(\ell)}\bcall{\set{f}}\in\L{2}\call{\set{X};\Reals[N\times{J_\ell}]}$ such that:
    \begin{align}
        \bracks{\vec{z}\bcall{\set{f},\set{h}}}_n &\coloneqq
        \inprod{\vecfun{h}_n,
            \op{H}^{(\ell+1)}\bcall{\sigma(
                \op{Z}_{\mat{J}}^{(\ell)}\bcall{\vecfun{f}_n}
            )}
        }
        \text{, and,}\\
        \bracks{\matfun{Z}_{\mat{J}}^{(\ell)}\bcall{\set{f}}}_n &\coloneqq
        \op{Z}_{\mat{J}}^{(\ell)}\bcall{\vecfun{f}_n}.
    \end{align}

    By definition, $\vec{z}\bcall{\set{f},\set{h}}$ conditioned on $\matfun{Z}_{\mat{J}}^{(\ell)}\bcall{\set{f}}$ follows a multivariate centered Gaussian distribution \(
        \mathcal{N}\parens[\big]{\vec{0},
            \mat{\Sigma}\parens[\big]{\matfun{Z}_{\mat{J}}^{(\ell)}\bcall{\set{f}}}
        }
    \) with covariance matrix:
    \begin{align}
        [
            \mat{\Sigma}\parens{
                \mat{A}
            }
        ]_{jk}
        = \angles[\big]{\vecfun{h}_k, \angles[\big]{\vecfun{h}_j,
            \fun{c}_{\op{H}^{(\ell+1)}}\bcall{
                \sigma(
                    [\mat{A}]_k
                )^\T
                \sigma(
                    [\mat{A}]_j
                )
                {}/{J_\ell}
            }
            \bm{I}_{J_{\ell+1}}
        }}.
    \end{align}
    
    Thus, by the tower rule, the characteristic function of the marginal distribution of $\vec{z}\bcall{\set{f},\set{h}}$ is:
    \begin{align}
        \phi_{\vec{z}\bcall{\set{f},\set{h}}}\call{\vec{t}} =
        \E\bcall{
            \E\bracksgiven{
                \exp\call{i\vec{t}^\T\vec{z}\bcall{\set{f},\set{h}}}
            }{\matfun{Z}_{\mat{J}}^{(\ell)}\bcall{\set{f}}}
        }
        = \E\bcall{\exp\call{
            \vec{t}^\T
            \mat{\Sigma}\parens[\big]{\matfun{Z}_{\mat{J}}^{(\ell)}\bcall{\set{f}}}
            \vec{t}
        }}.
    \end{align}

    Now, consider the point-wise convergence of the characteristic function:
    \begin{equation}
        \lim\limits_{J_{\ell-1}\to\infty}\cdots\lim\limits_{J_1\to\infty}
        \phi_{\vec{z}\bcall{\set{f},\set{h}}}\call{\vec{t}}
        \eqqcolon \phi_{\infty}\call{\vec{t}}
    \end{equation}
    Using the portmanteau theorem and continuity of $\mat{\Sigma}\call\blank$, we have that:
    \begin{align}
        \phi_{\infty}\call{\vec{t}}
        &= \E\bcall{\exp\call{
            \vec{t}^\T
            \mat{\Sigma}\parens[\Big]{
                \lim\limits_{J_{\ell-1}\to\infty}\cdots\lim\limits_{J_1\to\infty}
                \matfun{Z}_{\mat{J}}^{(\ell)}\bcall{\set{f}}
            }
            \vec{t}
        }}.
    \end{align}
    Now, our inductive hypothesis says that $\op{Z}^{(\ell)}_{\mat{J}}$ converges in distribution to a function-valued Gaussian process $\op{Z}^{(\ell)}_{\infty}$ with each output being i.i.d. With this fact, we can conclude that $\matfun{Z}_{\mat{J}}^{(\ell)}\bcall{\set{f}}$ also converges in distribution to the corresponding variable:
    \(
        \bracks{\matfun{Z}_{\infty}^{(\ell)}\bcall{\set{f}}}_n \coloneqq
        \op{Z}_{\infty}^{(\ell)}\bcall{\vecfun{f}_n}
    \). This means that:
    \begin{align}
        \phi_{\infty}\call{\vec{t}}
        &= \E\bcall{\exp\call{
            \vec{t}^\T
            \mat{\Sigma}\parens[\Big]{
                \matfun{Z}_{\infty}^{(\ell)}\bcall{\set{f}}
            }
            \vec{t}
        }},
    \end{align}
    which is the characteristic function of a variable defined as:
    \begin{align}
        \bracks{\tilde{\vec{z}}\bcall{\set{f},\set{h}}}_n &\coloneqq
        \inprod{\vecfun{h}_n,
            \op{H}^{(\ell+1)}\bcall{\sigma(
                \op{Z}_{\infty}^{(\ell)}\bcall{\vecfun{f}_n}
            )}
        }.
    \end{align}

    Therefore, $\vec{z}\bcall{\set{f},\set{h}}$ iteratively converges in distribution to $\tilde{\vec{z}}\bcall{\set{f},\set{h}}$, as $J_{\ell}\to\infty$ for every $\ell \leq \ell$. Since the set $(\set{f},\set{h})$ is arbitrary, we can conclude that \(
        (\op{H}^{(\ell+1)}\circ\sigma\circ\op{Z}_{\mat{J}}^{(\ell)})
    \) also converges in distribution to \(
        (\op{H}^{(\ell+1)}\circ\sigma\circ\op{Z}_{\infty}^{(\ell)})
    \) as a random operator.

    From the induction step, we know that the entries in $(\sigma\circ\op{Z}_{\infty}^{(\ell)})$ are i.i.d.\@ since the entries of $\sigma\circ\op{Z}_{\infty}^{(\ell)}$ are also i.i.d. Therefore, we use Lemma~\hyperref[app:sec:composition]{\ref*{lemma:composition}} to show that
    \(
        \lim\limits_{J_\ell\to\infty}
        (\op{H}^{(\ell+1)}\circ\sigma\circ\op{Z}_{\infty}^{(\ell)})
    \) converges in distribution to a function-valued Gaussian process with covariance function
    \begin{align}
        \matfun{C}^{(\ell+1)}\bcall{\vecfun{f}, \vecfun{g}}
        =
        \fun{c}_{\op{H}^{(\ell+1)}}\bracks[\big]{
            \fun{c}_{(\sigma\circ\op{Z}_{\infty}^{(\ell)})}
        }\bm{I}_{J_{\ell+1}}
        =
        \fun{c}^{(\ell+1|\ell)}\bracks[\big]{
            \fun{c}_{\sigma}\bracks[\big]{
                \fun{c}^{(\ell)}\bracks[\big]{\vecfun{f}, \vecfun{g}}
            }
        }\bm{I}_{J_{\ell+1}}.
    \end{align}

    Therefore, we just proved by induction that the iterated limit $\lim\limits_{J_{D-1}\to\infty}\cdots\lim\limits_{J_1\to\infty} \op{Z}_{\mat{J}}^{(D)}$ converges in distribution to a ${\op{Z}_{\infty}^{(D)} \follows \GPdist{\vec{0}, \fun{c}_{\infty}\mat{I}_{J_{\ell}}}}$ and this covariance function is equal to:
    \begin{equation}
        \fun{c}_\infty\bcall{\vecfun{f}, \vecfun{g}} =
        \fun{c}^{(d)}\bcall{\vecfun{f}, \vecfun{g}} =
        \fun{c}^{(d|d-1)}[
        \fun{c}^{(d-1|d-2)}[
        \cdots
        \fun{c}^{(2|1)}\bracks{
        \fun{c}^{(1)}\bracks{\vecfun{f}, \vecfun{g}}
        }\cdots]
    \end{equation}
\end{proof}

\section{Experimental details}
\label{ap:experiments}

In this section, we describe the setup for our experiments. As previously mentioned, all experiments were run in a desktop machine with a 3.8GHz Intel Core i7-9800X CPU and a 24GB NVIDIA Titan RTX (TU102) GPU. More details for each experiment can be found below.

\subsection{Empirical demonstration of results}

For both experiments, the input function $\fun{f} \colon \Torus \to \Reals$ has band-limit $B = 3$, with its output values $\fun{f}\call{x}$ sampled from a uniform distribution $\mathcal{U}(-1,1)$. In other words, we can express this band-limited function as:
\begin{align}
    \fun{f}\call{x}
    &=
    \frac{1}{7}
    \sum_{s=-3}^3
    f_s
    \sum_{s'=-3}^3
    \psi_{s'}\call{x-\frac{2\pi}{7}s},
\end{align}
where each $f_s\follows\mathcal{U}(-1,1)$ is independent and identically distributed.

In the first experiment of \cref{fig:variance_estimate}, we construct the operator layer $\op{H}$ under the usual formulation:
\begin{align}
    \op{H}\bcall{\fun{f}}\call{x}
    \colon \L{2}\call{\Torus}\to\L{2}\call{\Torus}
    = \op{A}_{\fun{k}}\bcall{\fun{f}}\call{x} + w\fun{f}\call{x},
\end{align}
where $w\follows\Normaldist{0,1}$ and $\fun{k}$ follows the band-limited Gaussian process distribution (\cref{sec:inf-fno,ap:sec:gp-bl}) with with band-limit $B=3$ and variance $\sigma^2 = 1/7$. Then, the operator on $\fun{f}$ is evaluated at zero $\op{H}\bcall{\fun{f}}\call{0}$ with increasing sample sizes.

For the second experiment of \cref{fig:fno_limit}, we construct the single-layer neural operator:
\begin{align}
    \op{Z}\bcall{\fun{f}}\call{x}
    \colon\L{2}\call{\Torus}\to\L{2}\call{\Torus;\Reals[J]}\to\L{2}\call{\Torus}
    = (\vec{w}_2^\T \circ \operatorname{ReLU} \circ{}\, \parens{\op{A}_{\vecfun{k}} + \vec{w}_1})\bcall{\fun{f}}\call{x},
\end{align}
where $J$ is the width of the hidden layer, and $\vec{w}_2\follows\Normaldist{0,1/J}$, $\vec{w}_2\follows\Normaldist{0,1}$, and $\vecfun{k}$ follows an i.i.d.\@ band-limited Gaussian process distribution (\cref{sec:inf-fno,ap:sec:gp-bl}) with band-limit $B=3$ and variance $\sigma^2 = 1/7$. For varying widths $J\in\braces{1,10,100,1000}$, we evaluate 10,000 samples of the operator on $\fun{f}$ at zero $\op{Z}\bcall{\fun{f}}\call{0}$ and show the density of the empirical distribution using kernel density estimation (KDE) with a Gaussian kernel.

These experiments are implemented in the file \texttt{experiments/fno\_limit.ipynb}.

\subsection{Regression}

We consider FNOs of increasing width, $J \in \braces{1, 10, 100}$ and $J \in \braces{1, 3, 10, 100, 500}$ for the synthetic and 1D Burgers' respectively, as well as $\infty$-FNOs, both with increasing kernel band-limits $B \in \braces{1, 5, 20}$. These single-layer neural operators are constructed as:
\begin{align}
    \op{Z}_{J,B}\bcall{\fun{f}}\call{x}
    \colon\L{2}\call{\Torus}\to\L{2}\call{\Torus;\Reals[J]}\to\L{2}\call{\Torus}
    = (\vec{w}_2^\T \circ \operatorname{ReLU} \circ{}\, \parens{\op{A}_{\vecfun{k}} + \vec{w}_1})\bcall{\fun{f}}\call{x},
\end{align}
where $J$ is the width of the hidden layer, and $\vec{w}_2\follows\Normaldist{0,1/J}$, $\vec{w}_2\follows\Normaldist{0,1}$, and $\vecfun{k}$ follow an i.i.d.\@ band-limited Gaussian process distribution (\cref{sec:inf-fno,ap:sec:gp-bl}) with variance $\sigma^2 = 1/(2B+1)$.

The hyperparameters of the $\infty$-FNO are estimated using L-BFGS, while the parameters of the FNOs are optimized with Adam using a step size of $0.001$. We evaluate all models using 5-fold cross-validation and report the average and standard deviation of the empirical $\L{2}$ norm of the prediction error. For $\infty$-FNOs, we use the posterior mean as the prediction.

This experiment is implemented in the file \texttt{experiments/train.py}.

\paragraph{Synthetic regression}

We start by defining the ground truth Fourier neural operator (FNO) which will generate our training and test data:
\begin{align}
    \op{Z}_{\opname{true}}\bcall{\fun{f}}\call{x}
    \colon\L{2}\call{\Torus}\to\L{2}\call{\Torus}\to\L{2}\call{\Torus}
    = ({w}_2 \circ \operatorname{ReLU} \circ{}\, \parens{\op{A}_{\fun{k}} + {w}_1})\bcall{\fun{f}}\call{x},
\end{align}
where the hidden layer's width is 1 and the band-limit of $\fun{k}$ is equal to 5. Next, we sample $n = 100$ input functions $\fun{f}_i \colon \Torus \to \Reals$ with the same band-limit $B = 5$ and uniformily-distributed outputs $\mathcal{U}(-1, 1)$, so that we have:
\begin{align}
    \fun{f}_i\call{x}
    &=
    \frac{1}{11}
    \sum_{s=-5}^5
    f_{is}
    \sum_{s'=-5}^5
    \psi_{s'}\call{x-\frac{2\pi}{11}s},
\end{align}
where each $f_{is}\follows\mathcal{U}(-1,1)$ is independent and identically distributed.
We then compute $\op{Z}_{\opname{true}}\bcall{\fun{f}_i}$ on an equally spaced grid given by $\braces{-5\frac{2\pi}{11}, \dotsc, 5\frac{2\pi}{11}} \subset \Reals[11]$.

\paragraph{1D Burgers' equation}

This dataset is provided from PDEBench \citep{PDEBench2022}, which includes solutions to the 1D Burgers' equation:
\begin{align}
 \frac{\partial}{\partial t}\fun{u}(t,x) + \frac{1}{2}\frac{\partial}{\partial x}\fun{u}^2(t,x) &= \frac{\nu}{\pi} \frac{\partial^2}{\partial x^2}\fun{u}(t,x),
\end{align}
where $x \in (0,1)$ and $t \in (0,2]$ are independent variables and $\nu$ is the diffusion coefficient. 

The regression task is set up with $\nu = 0.002$ and a collection of initial conditions $\braces{\fun{u}(0,\blank) = \fun{f}_i}_{i=1}^n$ and their respective end states $\braces{\fun{u}(2,\blank) = \fun{g}_i}_{i=1}^n$. Due to memory constraints when creating the covariance matrices for $\infty$-FNO, we subsample the original dataset to $n=100$ functions and a grid size of $m=103$. The original data can be downloaded at \url{https://darus.uni-stuttgart.de/api/access/datafile/268193}.